\documentclass[fleqn,10pt]{wlscirep}
\usepackage[utf8]{inputenc}
\usepackage[T1]{fontenc}
\usepackage{lineno}

\usepackage{hyperref}
\usepackage{xurl}
\usepackage[most]{tcolorbox}
\usepackage{pdfpages}

\title{Coordinated Pandemic Control with Large Language Model Agents as Policymaking Assistants}

\author[1,$\dag$]{Ziyi Shi}
\author[2,$\dag$]{Xusen Guo}
\author[1,*]{Hongliang Lu}
\author[2]{Mingxing Peng}
\author[2]{Haotian Wang}
\author[3]{Zheng Zhu}
\author[4]{Zhenning Li}
\author[2]{Yuxuan Liang}
\author[2]{Xinhu Zheng}
\author[1,*]{Hai Yang}

\affil[1]{The Hong Kong University of Science and Technology, Hong Kong}
\affil[2]{The Hong Kong University of Science and Technology (Guangzhou), China}
\affil[3]{Zhejiang University, China}
\affil[4]{University of Macau, Macau}

\affil[*]{corresponding authors: Hongliang Lu and Hai Yang}

\affil[$\dag$]{These authors contributed equally to this work}

\begin{abstract}
Effective pandemic control requires timely and coordinated policymaking across administrative regions that are intrinsically interdependent. However, human-driven responses are often fragmented and reactive, with policies formulated in isolation and adjusted only after outbreaks escalate, undermining proactive intervention and global pandemic mitigation. To address this challenge, here we propose a large language model (LLM) multi-agent policymaking framework that supports coordinated and proactive pandemic control across regions. Within our framework, each administrative region is assigned an LLM agent as an AI policymaking assistant. The agent reasons over region-specific epidemiological dynamics while communicating with other agents to account for cross-regional interdependencies. By integrating real-world data, a pandemic evolution simulator, and structured inter-agent communication, our framework enables agents to jointly explore counterfactual intervention scenarios and synthesize coordinated policy decisions through a closed-loop simulation process. We validate the proposed framework using state-level COVID-19 data from the United States between April and December 2020, together with real-world mobility records and observed policy interventions. Compared with real-world pandemic outcomes, our approach reduces cumulative infections and deaths by up to 63.7\% and 40.1\%, respectively, at the individual state level, and by 39.0\% and 27.0\%, respectively, when aggregated across states. These results demonstrate that LLM multi-agent systems can enable more effective pandemic control with coordinated policymaking. More broadly, this study presents a generalizable framework for operationalizing LLM agents in large-scale public policy settings, offering a promising decision-support paradigm for future pandemics and other complex societal challenges characterized by strong regional interdependence.
\\


\end{abstract}

\begin{document}

\flushbottom
\maketitle

\section*{Introduction}\label{sec1}

Pandemic control compels policymakers to make effective policies to safeguard public health. However, effective coordination across regions is intrinsically challenging, as many jurisdictions tend to prioritize short-term containment actions over coordinated, long-term planning \cite{asthana2025decision, holtz2020interdependence}. Such fragmented and myopic responses increase the difficulty of global pandemic control and, in some cases, may even exacerbate disease spread \cite{graff2020spread}. A key reason is regional heterogeneity. Pandemic situations vary widely across regions due to differences in population density, mobility patterns, healthcare capacity, socioeconomic structure, and data availability \cite{flaxman2020estimating, chang2020modelling}. Policymakers need to handle large volumes of heterogeneous information, often under severe time pressure, making it difficult to design both region-specific and globally coordinated interventions. Moreover, effective pandemic control necessitates proactive intervention under deep uncertainty. Transmission dynamics are multifactorial and evolve rapidly over time, meaning that delays in policy response can lead to irreversible public health consequences \cite{kissler2020projecting, prem2020effect}. Yet institutional constraints such as jurisdictional silos, delayed surveillance, and limited forecasting capacity often prevent governments from anticipating pandemic trends and acting preemptively \cite{domenici2022fragmented}. Even well-resourced governments have struggled to assess intervention trade-offs in advance and to deploy timely preventive measures \cite{sacco2023proactive}. Taken together, policymaking for pandemic control faces three fundamental challenges: 1) enabling effective coordination across regions, 2) accounting for region-specific and data-intensive heterogeneity, and 3) supporting proactive policy interventions under uncertainty.

The COVID-19 pandemic exemplified these challenges. During the global fight against COVID-19, the international response revealed substantial fragmentation across all levels of governance, from global coordination to local implementation \cite{domenici2022fragmented}. Many regions pursued policies independently; however, these siloed policies were often conflicting, leading to detrimental consequences for both public health and economic stability \cite{holtz2020interdependence}. This phenomenon is particularly pronounced in highly connected regions \cite{pacces2020diversity}. When neighboring regions adopted fragmented interventions, such as asynchronous lockdowns, uneven travel restrictions, or divergent reopening timelines, population mobility allowed the virus to continue circulating across administrative boundaries \cite{althouse2023unintended, lu2023understanding, van2025four, perofsky2024impacts}. As a result, stringent control measures implemented by individual regions were frequently undermined by inflows of infections from less-restricted neighbors, rendering localized efforts partially ineffective. Modeling studies of the United States have demonstrated that such uncoordinated interventions led to significant efficiency losses and worsened overall outcomes \cite{holtz2020interdependence}. 
Similar patterns were also observed in Europe \cite{pacces2020diversity, ruktanonchai2020assessing}. As concluded by the Lancet Commission, “\textit{widespread, global failures at multiple levels in the COVID-19 response led to millions of preventable deaths}” \cite{sachs2022lancet}. A critical lesson is highlighted: it is difficult for purely human-driven policymaking to achieve effective coordination and real-time response; what often follows are miscommunication, policy misalignment, and suboptimal public health outcomes \cite{graff2020spread, flaxman2020estimating}.

In response to these challenges, policymakers have resorted to established response paradigms and local heuristics. Intervention packages are typically designed based on past influenza models, community norms, and classical epidemiological strategies, such as the influenza control framework proposed by Ferguson et al.\ \cite{ferguson2006strategies}. However, real-world implementation remains piecemeal and reactive, largely due to fragmented governance structures, delayed situational awareness, and limited capacity to synthesize heterogeneous data into timely, coordinated policy decisions.
At both national and regional levels, policy actions were frequently triggered by case surges or political pressure rather than informed by forward-looking analysis \cite{mckee2025power, bicher2022supporting}. Converging evidence suggests that proactive strategies would have substantially outperformed reactive ones. For example, Sacco et al.\ showed that jurisdictions adopting preemptive interventions exhibited far greater resilience than those that delayed action until outbreaks intensified \cite{sacco2023proactive}. Yet effective decision-support tools and institutional agility remain limited, leaving pandemic policymaking heavily dependent on static “plans on paper” and expert judgment. Even where complex epidemiological models were available, their influence on decision-making was often constrained. In a few cases, such as Austria’s national modeling consortium, ensembles of epidemiological forecasts were used to inform the timing of policy interventions \cite{bicher2022supporting}. More commonly, however, officials defaulted to simple rules of thumb, resulting in a patchwork of measures shaped by local political considerations rather than systematic optimization. Therefore, an effective paradigm is urgently needed to transform policymaking from a siloed, reactive process into a more coordinated and proactive one \cite{prem2020effect,kissler2020projecting}.

Recent advances in large language models (LLMs) have demonstrated their outstanding ability to handle heterogeneous information, perform natural language reasoning, and adapt to complex and evolving contexts \cite{zhao2023survey}. Compared with conventional policy support methods based on fixed rules or pre-established optimization objectives, LLMs offer flexible, context-aware reasoning capabilities, making them promising tools for interpreting ever-changing pandemic situations. Existing studies have explored LLM applications in disease forecasting \cite{saeed2024llm4cast, du2025advancing}, pandemic modeling and simulation \cite{kwok2024utilizing, williams2023epidemic}, and public health information analysis \cite{zhou2023traditional, consoli2024epidemic}. However, these works primarily treat LLMs as predictive models or heterogeneous information processors, lacking mechanisms for interaction with dynamic environments or decision adaptation through feedback. Consequently, they fall short of addressing the sequential, interactive, and coordination-intensive demands of real-world pandemic control. Against this backdrop, the recent emergence of LLM agents opens a new window. LLM agents operationalize language models as autonomous decision-making entities that maintain internal state and interact with environments or other agents \cite{guo2024large, chopra2024limits}. This agent-based paradigm enables LLMs to support sequential decision-making, adapt to evolving system dynamics, and reason about coordination among multiple stakeholders. Such capabilities align naturally with coordinated pandemic control, where policies must be continuously made under uncertainty and coordinated across regions and administrative levels. Given the prowess of LLM agents, a compelling question arises: if we had LLM agents as our policymaking assistants during past pandemics, could more effective pandemic control have been achieved?

\begin{figure}[htbp]
  \centering
  \includegraphics[trim={0.9cm 0cm 0.9cm 0cm}, clip, width=\columnwidth]{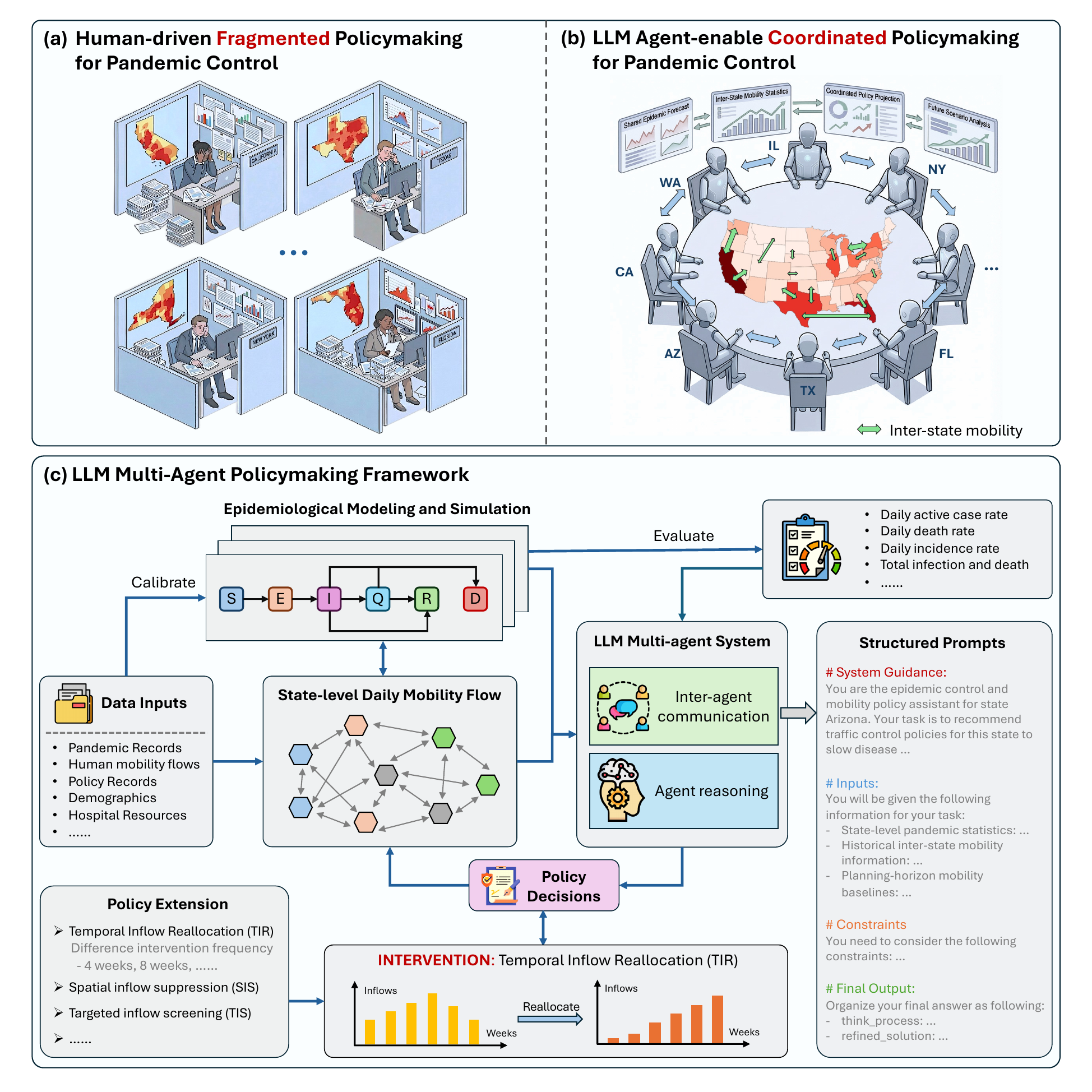}
  \caption{Comparison between traditional pandemic policymaking and our proposed LLM-agent-enabled coordinated policymaking. \textbf{a}, Traditional pandemic policymaking under fragmented coordination. Regional authorities operate largely in isolation, relying on delayed or incomplete information and independent analyses. Limited cross-region communication and asynchronous interventions lead to misaligned policies and inefficient pandemic control. \textbf{b}, LLM multi-agent system for coordinated pandemic policymaking. Each administrative region is represented by an LLM agent that exchanges information with others, enabling collaborative reasoning and coordinated policy decisions across regions while accounting for inter-state mobility. \textbf{c}, Architecture of the proposed LLM-agent system for pandemic policymaking. Each regional agent integrates epidemiological modeling, real-world data inputs (case records, mobility flows, demographics, and healthcare resources), and structured prompts to reason about policy actions. Inter-regional mobility is modeled through state-level mobility flows, inducing cross-region transmission dynamics. Agents communicate, evaluate outcomes, and generate coordinated policy decisions, which are instantiated as intervention strategies (e.g., temporal inflow reallocation, spatial inflow suppression, and targeted inbound screening) and fed back into the pandemic simulation loop.}
  \label{fig:fig1}
\end{figure}

In this study, we develop an LLM multi-agent policymaking framework for coordinated pandemic control, in which each LLM agent serves as a regional AI assistant to support proactive and coordinated policymaking across administrative units. As illustrated in Fig.~\ref{fig:fig1}, our framework follows a three-step closed-loop decision process. 
First, for each region, the LLM agent calibrates a pandemic transmission model using observed epidemiological and mobility data, and uses it to simulate short-term pandemic evolution. This step captures how inter-regional mobility shapes the pandemic situation, as reflected in some key indicators like active case counts and mortality.
Second, these pandemic indicators and mobility flow data are fed into the LLM multi-agent policymaking system. Through inter-agent communication and collective reasoning, these agents exchange information and jointly determine policy actions designed to control disease spread by adjusting inter-regional mobility. 
Finally, the policies are applied to update inter-regional mobility flows, which are fed back into the pandemic transmission model to simulate the pandemic evolution under the new policy settings. The updated pandemic situations subsequently serve as inputs to the next decision cycle, enabling iterative and anticipatory policy refinement.
We validate the proposed framework using real-world COVID-19 data from the United States between April and December 2020, together with high-resolution mobility records and real-world policies.
Compared to real-world pandemic situations, our LLM-agent system, 1) at the individual-state level, reduces cumulative infections and deaths by up to 63.7\% and 40.1\%, respectively; and 2) at the aggregate level across states, reduces cumulative infections and deaths by 39.0\% and 27.0\%, respectively.
Looking ahead, this study showcases a generalizable framework for operationalizing LLM agents into large-scale public policy systems, suggesting a promising decision-support paradigm for future pandemics and other complex societal crises that require cross-regional coordination.

\section*{Results}\label{sec2}

\subsection*{LLM Multi-Agent Policymaking Framework}

To facilitate coordinated pandemic control among multiple states, we develop a multi-agent policymaking framework. Below, we illustrate how the framework operates.

\textbf{Formulation:} In our framework, each state is assigned an LLM agent as the policymaking assistant, which leverages real-time pandemic information from its own state as well as other states to dynamically manage local inbound mobility. Central to our policymaking is the Temporal Inflow Reallocation (TIR) protocol, under which the total inbound volume from each origin state is held constant over a prescribed planning horizon (e.g., eight weeks). Accordingly, the study period is divided into multiple reallocation cycles. In each cycle, the agents proactively redistribute weekly quotas to modulate the influx rate for their origin states. To illustrate, if State A’s total inbound flow from State B is 80,000 over 8 weeks, TIR does not alter this 80,000-person aggregate; instead, it would reallocate the distribution from a uniform 10,000 per week to a non-linear schedule (e.g., 5,000 in week 1 and 15,000 in week 4) based on real-time pandemic situations. This design ensures that our results remain grounded in realistic mobility constraints, while enabling the exploration of potentially more effective pandemic control policies.

\textbf{Data:} We leverage real-world COVID-19 data from the United States spanning April to December 2020. To consider multi-source and heterogeneous pandemic information, we incorporate and align three categories of data:
1) state-level daily pandemic records across the United States;
2) state-to-state daily human mobility flows; and
3) state-level policy records.

\textbf{Simulation:} 
We conduct simulation experiments based on real-world pandemic and mobility data spanning from April 12 to December 31, 2020. 
Once interstate flows are regulated by LLM-agent-generated policies, the SEIQRD model-based epidemiological simulator evolves the pandemic dynamics accordingly \cite{wang2018inferring,balcan2009multiscale}, as reflected in variations in state-level changes in infection and mortality. 

To provide a detailed policy analysis, a 5-state experiment is first conducted among Arizona (AZ), Mississippi (MS), New Mexico (NM), Texas (TX), and Virginia (VA).
We then verify our framework on 20 states to examine its scalability.
A detailed formulation is provided in the Methods section.

\textbf{Prompt:} 
Statistics observed during the simulation are collected at each policymaking cycle and standardized into structured textual inputs for prompt construction.
Each prompt comprises three components: 1) system guidance, which defines the objective for mitigating pandemic spread and the control strategy (e.g., TIR, see Appendix for prompt example);
2) state-level pandemic statistics, which capture the spatio-temporal epidemiological trajectories within the local state and its neighboring states (e.g., daily infection counts and mortality rate); and 
3) mobility information, which integrates historical inter-state mobility patterns and anticipated inbound mobility flow from other states over the planning horizon.

Ultimately, the framework outputs a set of reallocation proportions, which specify how the total inflow is partitioned over the planning horizon.

\textbf{Evaluation:} 
Cumulative metrics and temporal indicators, obtained from the simulation,  are used for evaluation. 
The cumulative metrics, including total infections and mortality, serve as an aggregate measure of long-term policy efficacy.
Regarding temporal indicators, we monitor the daily incidence rate (IR), death rate (DR), and active case rate (ACR) for each state.
In doing so, we can evaluate both localized performance at the state level while quantifying the emergent systemic impact at the aggregate level.

\textbf{Validation:} 
We consider and compare four distinct policymaking paradigms in our validation:
1) LLM-agent-based policymaking, which performs coordinated policymaking based on system dynamics;
2) ground-truth policymaking, reflecting real-world, human-led pandemic control measures;
3) expert-guided policymaking, in which heuristic travel control strategies are designed based on changes in pandemic statistics; and
4) random policymaking, which adjusts mobility in the absence of explicit optimization objectives.

\textbf{Extension:} To test the extensibility of our framework, we account for a diverse set of policy designs. In particular, we evaluate policy outcomes under varying policymaking frequencies (i.e., reallocation cycle lengths), including bi-monthly (8-week TIR) and monthly (4-week TIR) planning horizons, as well as different intervention strategies such as direct mobility suppression (i.e., spatial inflow suppression, SIS) and targeted inbound screening (TIS).

\subsection*{5-state Results of Policy Evaluation and Interpretation}

\begin{figure}[t]
  \centering
  \includegraphics[width=\columnwidth]{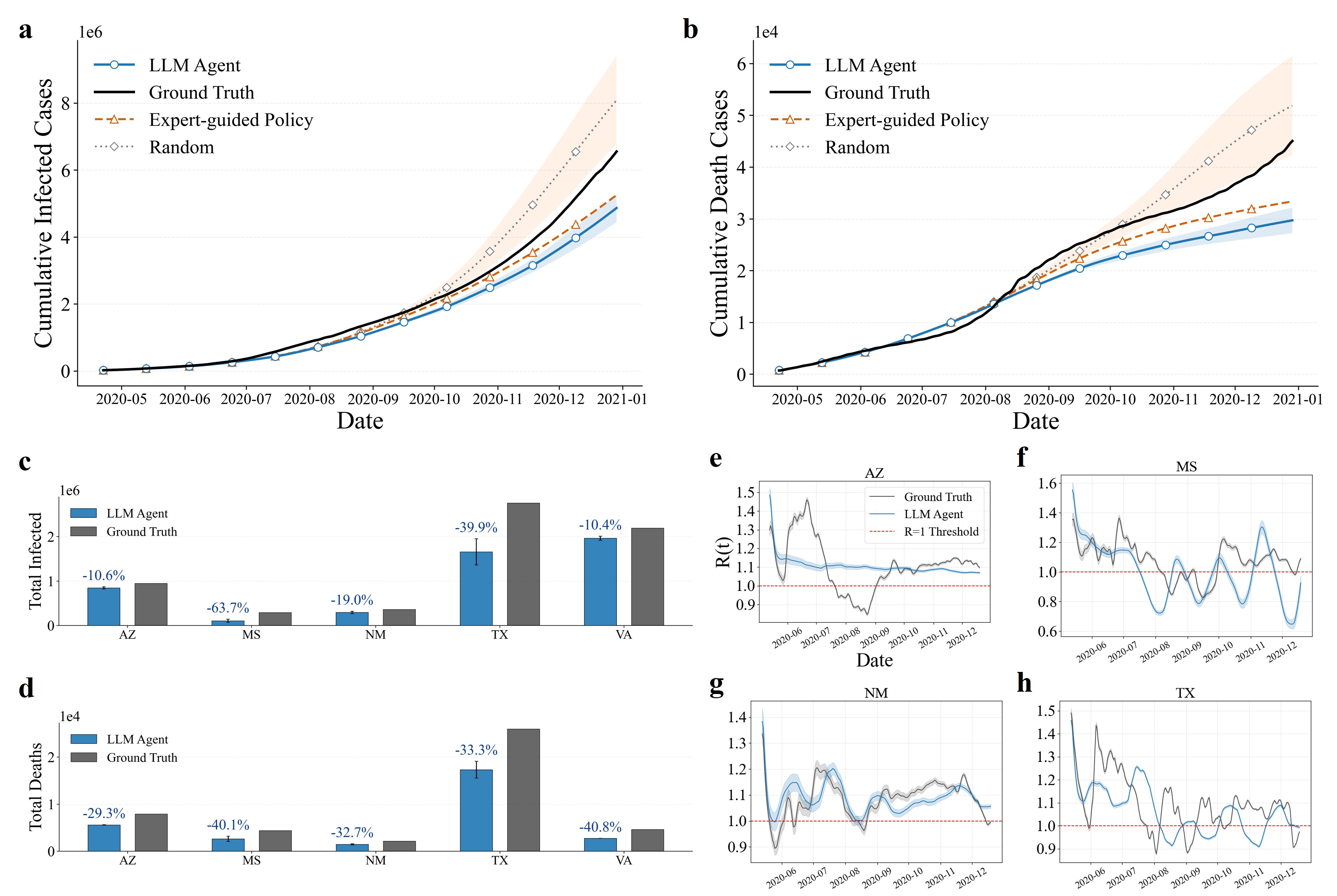}
  \caption{\textbf{Effectiveness of the LLM multi-agent policymaking framework in mitigating pandemic spread.} 
  The cumulative confirmed infections \textbf{(a)} and cumulative deaths \textbf{(b)} vary across the time (from May 2020 to December 2020). Results are compared across different intervention strategies, including the LLM agent–based policymaking, expert-experience-based policymaking,  random policymaking, and the observed ground-truth policymaking. 
  \textbf{c},\textbf{d}, The overall reduction in cumulative infections and deaths achieved by the LLM agent–based policy differ across five states, where numerical annotations on each bar denote the percentage reduction in cumulative infections and mortality relative to the ground truth. Furthermore, \textbf{e-h}, LLM agent-based policy stabilizes the temporal dynamics of effective reproduction number $R_t$ and reduces its mean value.}
  \label{fig:fig2}
\end{figure}

In this section, we evaluate the proposed LLM multi-agent policymaking framework on a 5-state setting, comparing its performance against the ground-truth, expert-guided, and random policymaking paradigms, with the planning horizon set to 6 weeks by default.
We then analyze the interpretability of policy outcomes from both qualitative and quantitative perspectives.

Fig.~\ref{fig:fig2}a–d show the results of the proposed framework in pandemic control at aggregate- and state-level. 
First, at the aggregate level, the policy generated by LLM agents reduces the total cumulative infected and death cases by 25.7\% and 34.0\% compared with the ground truth. 
The random policy performs worse than the ground-truth policy due to its lack of a consistent objective.
Although the expert-guided policy controls the spread of the pandemic, the LLM agent policy further improves overall performance.
These results underscore both the importance of cross-state coordination and the challenges of achieving it when individual states pursue self-interested policies.
Second, at the state level, the LLM agent policy exhibits positive but heterogeneous levels of effectiveness across states.
In states with less restrictive ground-truth control policies, such as MS and TX, performance improvements are more pronounced (from 33.3\% to 63.7\%). 
For states where ground-truth measures are strict, such as AZ and VA, the reduction in infection counts remains modest (approximately 10\%), whereas the reduction in mortality is substantial (29.3\% and 40.8\%, respectively).
Moreover, Fig.~\ref{fig:fig2}e-\ref{fig:fig2}h present the variation of effective reproduction number $R_t$ across states, characterizing the pandemic transmissibility (See "Method" Section for the calculation of $R_t$). 
The LLM agent policy mitigates transmissibility by lowering and stabilizing $R_t$.
However, $R_t$ remains above the threshold of one, indicating that an infected individual generates more than one case on average.
So, transmission is slowed but not fully suppressed.

\begin{figure}[htbp]
  \centering
  \includegraphics[width=\columnwidth]{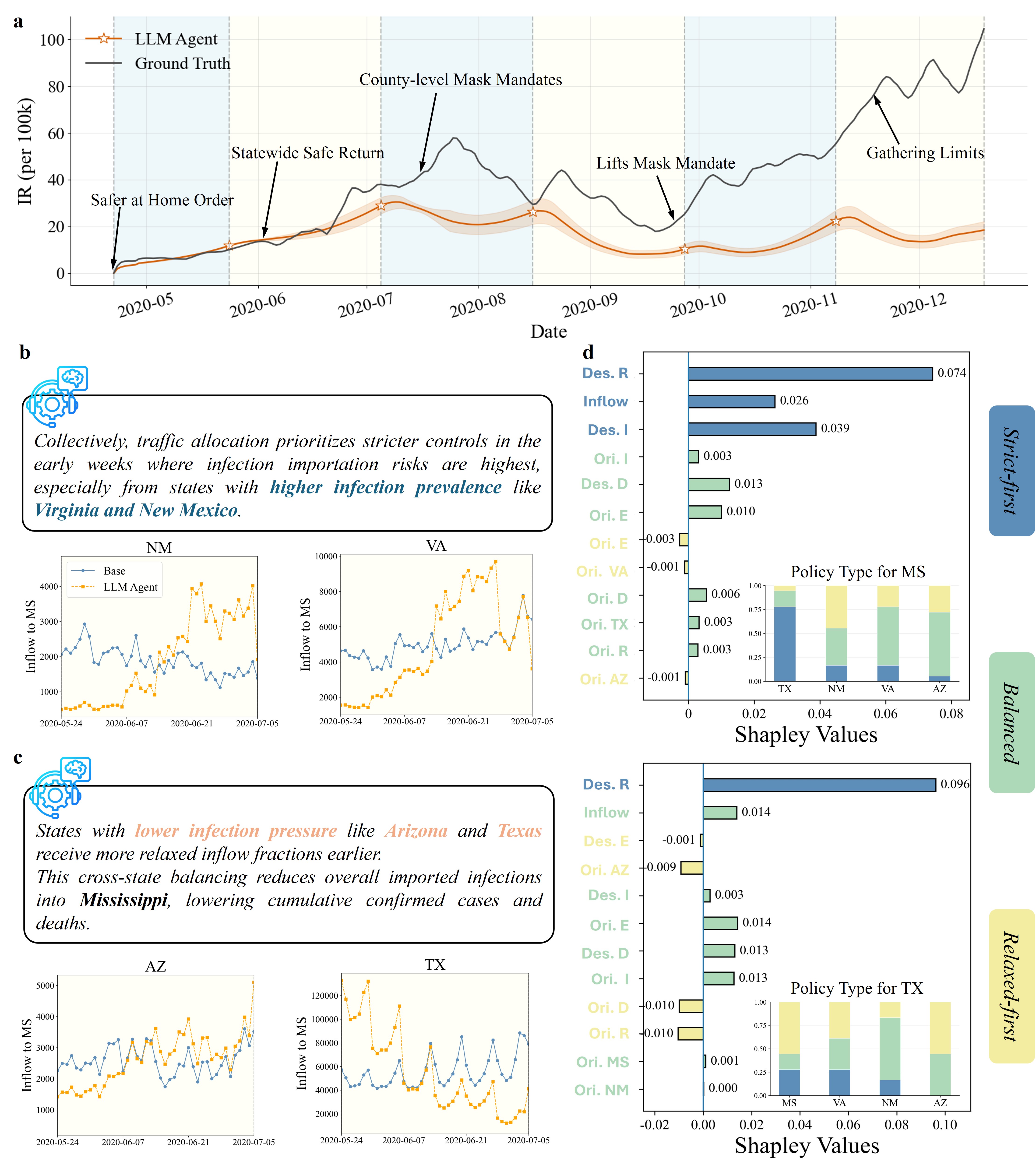}
  \caption{\textbf{State-level intervention illustration and interpretation.} \textbf{a}, The real-world policy is compared with the mobility reallocation policy generated by the LLM agent in MS (Mississippi), together with the corresponding daily incidence rate. Each reallocation cycle is highlighted in the background using alternating light blue and yellow shading. \textbf{b},\textbf{c}, The specific mobility adjustments during the second reallocation cycle and the reasoning process of LLM agents are illustrated. Moreover, \textbf{d}, LLM agent policy are classified into three types (strict-first, balanced, and relaxed-first). Policy-type distributions for Mississippi and Texas are presented, along with each feature’s marginal contribution (i.e., the Shapley Values)  on strict-first policy.
}
  \label{fig:fig3}
\end{figure}

To illustrate how LLM agents operate in policymaking, we select Mississippi (MS) as an example and compare the LLM agent policy with the observed real-world one.
Fig.~\ref{fig:fig3}a shows the comparison results in IR.
The implementation of the LLM agent policy is marked by orange five-pointed stars, which divide the experimental period into six successive reallocation cycles. 
Under the LLM agent policy, IR is maintained below 40, whereas under the ground-truth policymaking, IR rises to nearly 60 by the end of July 2020 and continues to increase over 80 after November 2020.
This divergence emerges after June 2020, following the issuance of the “Statewide Safe Return” policy in reality. 
Thereafter, despite the implementation of mask mandates and limits on gatherings, pandemic transmission accelerated and became out of control.
To detail LLM agent policy during the same period, we visualize the state-to-state inflow to MS before and after the adjustment at the second reallocation cycle in Fig.~\ref{fig:fig3}b and \ref{fig:fig3}c, alongside the LLM reasoning process for qualitative explanations.
The reasoning result shows that LLM agents can identify high-risk origin states (i.e., NM and VA) and impose stricter early-stage interventions, resulting in inbound flow quotas below the uniform level.
This strategy limits the early imported infections from high-risk origins, thereby reducing the initial incidence rate and preventing subsequent exponential amplification of transmission.
For states with lower infection pressure (AZ and TX), they have more relaxed inflow fractions in the early weeks. 
Under a fixed quota, this approach constrains subsequent mobility and mitigates pandemic propagation driven by later infection surges.

Next, we provide a quantitative interpretation of LLM agent policymaking.
Based on the adjusted inflow distributions, state-level policies at each reallocation cycle are classified into three types: strict-first, relaxed-first, and balanced, corresponding to different pandemic phases and risk levels across states (see the Methods section for detailed classification criteria).
Their distribution and feature attribution for states MS and TX are illustrated in Fig.~\ref{fig:fig3}d.
Policymaking in MS prioritizes a strict-first intervention on TS, because TS exhibits the fastest growth in infections, while applying a balanced control strategy to the other three states.
As for Policymaking in TX, the relaxed-first policy is frequently applied to the origin states MS and AZ.
This policy decision indicates that MS and AZ pose a non-negligible but manageable importation risk to TX, for which gradual restrictions are effective to curb transmission.

To further quantify the impact of state-level features on the choice of policy types, we use the Shapley value to estimate each feature’s average marginal contribution to policymaking across all feature combinations. 
We compile epidemiological statistics—including recovered, exposed, infected, and deceased counts—for both origin and destination states, together with inflow volumes and a one-hot representation of the origin state.
The Shapley values of these features for "strict-first" policy are illustrated in Fig.~\ref{fig:fig3}d, and key findings are summarized.
First, for MS and TX, the recovery count in the destination state contributes largely to the strict-first policy (Shapley values 0.074 and 0.096, respectively), followed by inflow volume (0.026 and 0.014, respectively).
This result suggests that the agent imposes early stringent interventions when the local state has experienced extensive past transmission and faces elevated inflow pressure.
Second, infection counts in both the origin and destination states are positively associated with the adoption of the strict-first policy. 
In contrast, higher exposure levels (at the origin for MS and destination for TX) reduce the likelihood of adopting a strict-first policy.
These conditions, which signal potential infection growth in subsequent weeks, call for a gradual tightening of interventions.
Moreover, in TX, origin-level recoveries and deaths exhibit negative Shapley values for the adoption of the strict-first policy, whereas such effects are not observed in MS. 
This asymmetry reflects TX’s role as a high-mobility hub, where rapid population movements render current infection levels (i.e., infected and exposed cases) more informative for early intervention than historical pandemic levels (recoveries and deaths).

\begin{figure}[h]
  \centering
  \includegraphics[width=\columnwidth]{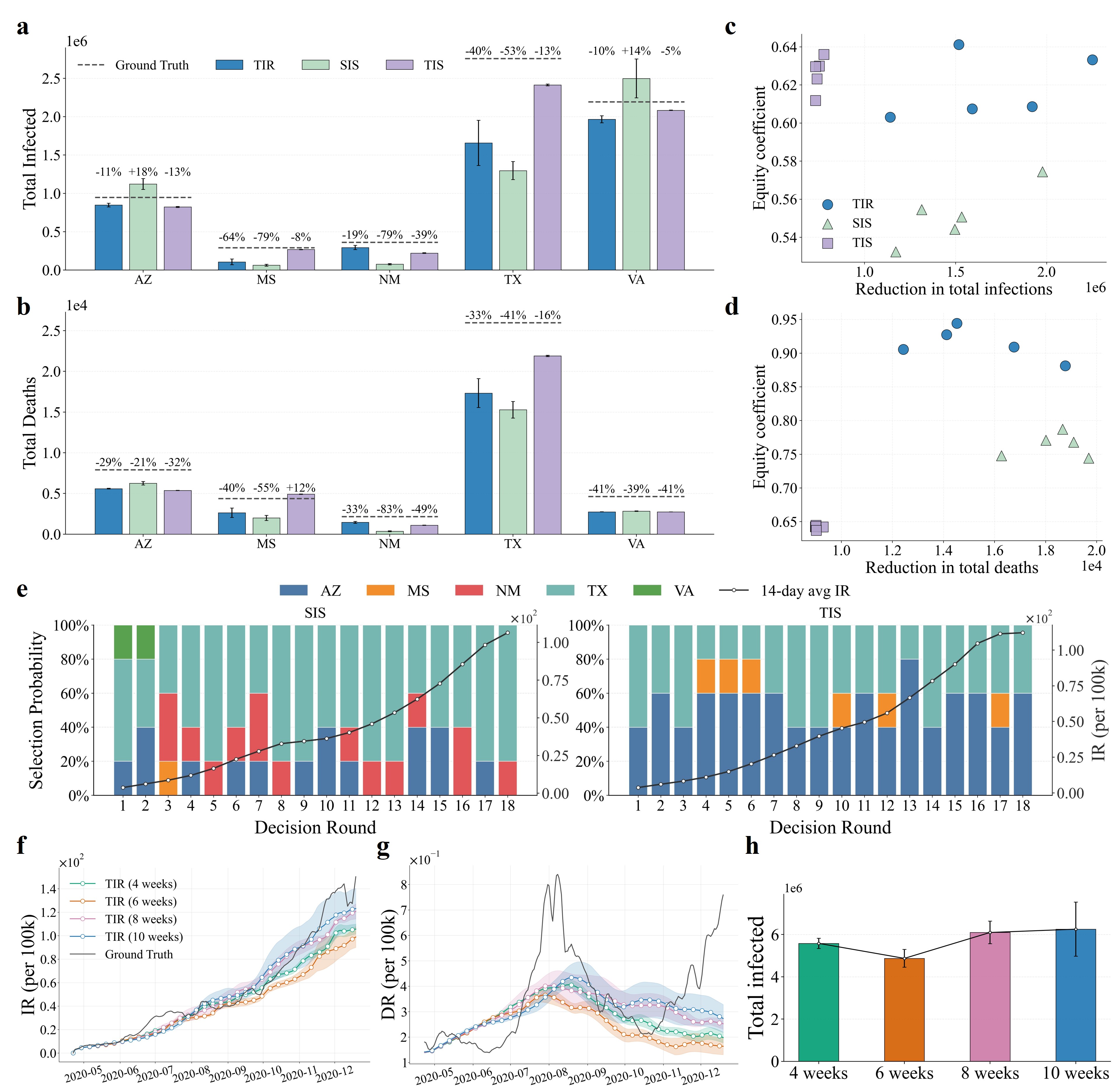}
  \caption{\textbf{Performance comparison across multi-dimensional policies.} 
First, \textbf{a},\textbf{b}, the state-level performance of different intervention strategies (TIR, SIS, and TIS) in terms of total infections and deaths is illustrated. The real-world conditions are indicated by black dashed lines, with the corresponding increase or reduction percentages under different intervention strategies annotated above.
Then, \textbf{c},\textbf{d}, the system-level indicators, equity coefficient and the total reductions in infections and deaths, are compared across strategies.
Each strategy is evaluated over five runs, with each point representing a single run.
\textbf{e}, The selection probabilities of different states under SIS and TIS across the study period are presented.
Moreover, \textbf{f}–\textbf{h}, TIR under planning horizons of 4, 6, 8, and 10 weeks is compared, evaluated by incidence rate (IR), death rate (DR), and total infections. 
   }
  \label{fig:fig4}
\end{figure}

\subsection*{5-state Results of Policy Extension}

Beyond TIR, we extend the proposed policymaking framework to additional intervention strategies for comparison.
Specifically, we consider two alternatives: Spatial Inflow Suppression (SIS) and Targeted Inbound Screening (TIS).
For SIS, each state identifies a specific high-risk origin state to reduce its inbound traffic flow by half over a two-week window, with policy decisions updated biweekly throughout the study period.
While this strategy provides immediate localized mitigation for the destination state, it also induces a spatial spillover effect, where suppressed traveler flows are redistributed to alternative, non-restricted states.
Under TIS, each state implements screening for travelers from a specific origin state over the subsequent two-week period, with policy decisions updated on the same biweekly schedule. This screening process identifies asymptomatic or presymptomatic individuals (E and I compartments) and transfers them into the confirmed (Q) compartment, thereby neutralizing their potential for onward transmission.
In both SIS and TIS, each decision round targets only one origin state to avoid excessive simultaneous restrictions.

Most states experience reductions in total infections and deaths under different intervention strategies, as shown in Fig.~\ref{fig:fig4}a and \ref{fig:fig4}b.
Taking TX, the most severely affected state, as an example, total deaths are reduced by 33\%, 41\%, and 16\% under TIR, SIS, and TIS, respectively.
Meanwhile, the three intervention strategies exhibit distinct effect patterns across states.
1) For SIS, although infections are reduced by more than 50\% in MS, NM, and TX, pandemic spread is even exacerbated in AZ and VA, with more than 10\% infection.
This effect is induced by the spatially redistributed interstate flows: mobility suppressed from high-risk origins is diverted to alternative, non-restricted states, unintentionally increasing their exposure and amplifying local transmission.
2) TIS exhibits weaker but more stable effects than SIS and fails to reduce mortality in MS.
3) Under TIR, infection and death counts decrease across all five states, achieving the most stable performance among the intervention strategies.

Fig.~\ref{fig:fig4}c and \ref{fig:fig4}d show system-level trade-offs between effectiveness (i.e., reduction in total infection or deaths) and equity (i.e., dispersion of improvements among states), where each point corresponds to a single run of the strategy.
A higher equity coefficient indicates a more evenly distributed improvement (see Methods for detailed calculation of equity coefficient).
The main findings are summarized as follows:
First, for both infection and death outcomes, TIR largely dominates the Pareto frontier, indicating its overall superior capability in balancing equity and efficiency in pandemic control. 
Second, with respect to infection control, TIS exhibits relatively high equity but limited effectiveness, whereas SIS achieves lower equity. 
Third, SIS substantially reduces total deaths, albeit with lower equity compared to TIR. It is mainly because SIS induces inter-agent competition through spatial flow redistribution, whereas TIR mitigates such competition.

To further examine inter-agent competition, the detailed policies under SIS and TIS are illustrated in Fig.~\ref{fig:fig4}e.
Specifically, the bar charts report, for each decision round, the proportion of times each state is selected by other destination states as the intervention state under SIS or TIS.
Under TIS, selection probabilities remain stable throughout the study period, with AZ and TX being the primary targets for screening.
By contrast, under SIS, when a particular high-risk state is identified by all other states, coordination is effectively enforced, thereby eliminating flow spillover effects. However, such coordination is not always achieved, resulting in diverted flows that surge into other states (e.g., in rounds 4 and 6, not all states suppress flows from TX).
As a result, SIS and TIS exhibit comparable IR, as the spillover-induced amplification under SIS offsets its localized suppression benefits.

Finally, we examine how the policymaking frequency of LLM agents influences the effectiveness of pandemic control. 
Different planning-horizon lengths (4, 6, 8, and 10 weeks) are compared under TIR,  with the corresponding IR, DR, and total infections shown in Fig.~\ref{fig:fig4}f–\ref{fig:fig4}h.
Overall, results across these strategies follow the ground-truth trend while attenuating transmission intensity and mortality to varying extents.
Notably, total infections under the 6-week TIR setting are reduced by more than 8\% relative to the 4-week and 8-week TIR settings.
This finding suggests that planning horizons that are either too long or too short lead to inferior performance.
The detailed state-level comparison results are presented in the Supplementary Figs 1-15.

\subsection*{20-state Results of Scalability}

To further evaluate the scalability of the proposed LLM multi-agent policymaking framework, we extend the analysis to a large scale.
After excluding states with incomplete statistical records, a total of U.S. 20 states are included in the experiment: Alabama (AL), Arizona (AZ), Arkansas (AR), Idaho (ID), Indiana (IN), Iowa (IA), Kentucky (KY), Michigan (MI), Minnesota (MN), Mississippi (MS), Nebraska (NE), New Mexico (NM), Ohio (OH), Oklahoma (OK), South Carolina (SC), Tennessee (TN), Texas (TX), Utah (UT), Virginia (VA), and Wisconsin (WI).
By comparing state-level transmissibility with cumulative infections and deaths, we examine the performance of the proposed framework in large-scale pandemic control.

\begin{figure}[!h]
  \centering
  \includegraphics[width=\columnwidth]{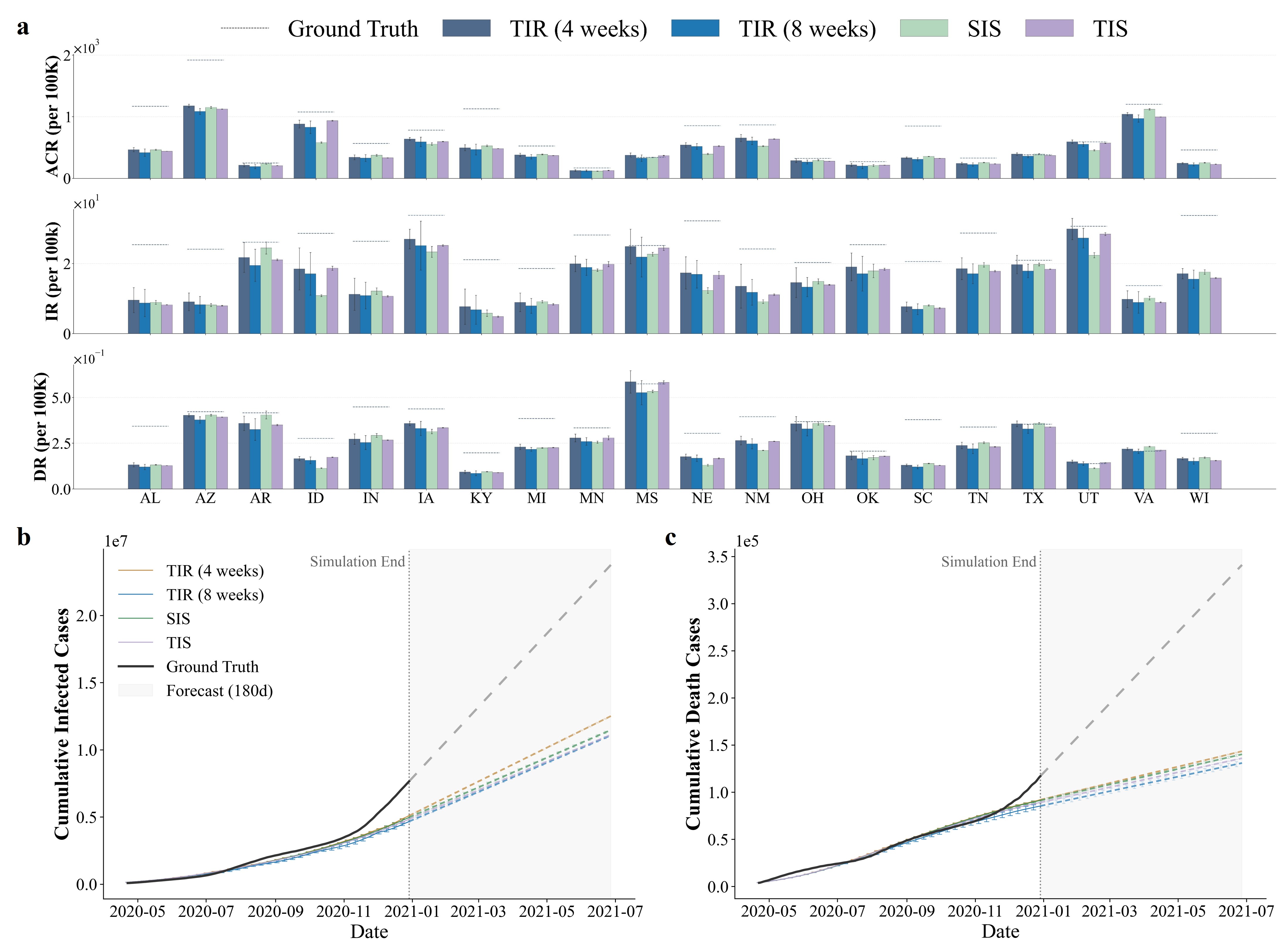}
  \caption{\textbf{Pandemic control performance in 20-state experiments.} \textbf{a}, State-level comparisons of the average daily active case rate (ACR), daily incidence rate (IR), and daily death rate (DR) across 20 states demonstrate the generalizability of the proposed framework. 
 \textbf{b}, Total cumulative infected cases under the LLM agent–based policy are compared with the ground-truth outcomes over the study period. 
In addition, we project the evolution of cumulative infections over the subsequent 180 days under this intervention pattern to demonstrate the long-term performance.}
  \label{fig:fig5}
\end{figure}

At the state level, infection and mortality rate under LLM agent policy are lower than those observed under ground-truth policymaking. 
The IR, DR, and ACR with different policy designs (4-week TIR, 8-week TIR, SIS, and TIS) across states are illustrated in Fig.~\ref{fig:fig5}a.
LLM agent policies lead to reductions in the ACR in almost all states, indicating suppression of active cases and an alleviation of healthcare system burdens.
A similar downward pattern is observed for IR and DR, indicating reductions in transmissibility and mortality.
By comparison, 8-week TIR delivers more stable and effective outcomes than 4-week TIR, especially in AL, AZ, AR, and IN.
As noted earlier, SIS exhibits greater variability in its state-level effects.
In some states, such as ID and NE, SIS achieves stronger outcomes than 8-week TIR, whereas in others (e.g., AR and OH) this strategy results in the weakest improvement among strategies.
TIS exhibits relatively stable effects across states; however, because this strategy does not alter mobility volumes, its impact remains modest in some states like MS and TX.

At the aggregate level, the LLM multi-agent policymaking framework mitigates the pandemic burden by reducing total infection and death counts.
By the end of 2020, cumulative infections and deaths across the 20 states can be reduced by more than $2.6 \times 10^{6}$ and $2.5 \times 10^{4}$, respectively.
Fig.~\ref{fig:fig5}b illustrates the temporal evolution of aggregate infections and deaths under different policy designs, with distinct colors representing alternative intervention trajectories.
All four tested policymaking strategies outperform the ground-truth policy, with reductions becoming pronounced after September 2020.
By comparison, 8-week TIR achieves the strongest improvement, followed by TIS. 
Notably, even the least effective intervention yields substantial system-level benefits, establishing a conservative lower bound on the effectiveness of coordinated policymaking at scale.
To further demonstrate the control effects, we forecast cumulative infections and deaths over the subsequent 180 days under different policymaking.
By that time, cumulative infections are reduced by approximately 47–53\% relative to the ground-truth policy, while cumulative deaths decline by 58–62\%.
The detailed state-level results for total infection and deaths are listed in Supplementary Figs 29-48. 
These results underscore the sustained and long-term effectiveness of LLM-agent–enabled policymaking in mitigating pandemic outcomes.

\section*{Discussion}\label{sec3}

Our work introduces an LLM multi-agent framework for coordinated pandemic policymaking, in which LLM agents serve as regional policy assistants that collaborate across administrative boundaries. In contrast to traditional approaches that rely on static rules or siloed policymaking~\cite{berger2021rational}, the proposed framework enables coordinated policymaking by allowing agents to share regional pandemic situations and jointly make policies that regulate cross-region mobility over time. Empirical validation using COVID-19 data shows that this coordination-centric paradigm improves pandemic control compared with ground-truth, human-driven responses. Moreover, our framework is scalable and generalizable: it can be deployed at the national scale by instantiating each administrative unit as an autonomous agent, and beyond SEIQRD, it can accommodate diverse policy interventions by incorporating different domain-specific simulators, without altering the underlying coordination and reasoning mechanisms. Together, these properties demonstrate the potential of LLM multi-agent systems for assisting real-world pandemic policymaking, particularly in mitigating the fragmentation and reactivity that characterize many conventional approaches.

Considering the diversity of LLMs, we evaluate the proposed framework using multiple LLM backbones while keeping the agent architecture and policy settings fixed. As shown in extended Fig.~\ref{fig:fig7} in Appendix~\ref{appendix:llm_comparison}, the framework exhibits consistent improvement across the tested models relative to ground-truth policymaking, although the magnitude and stability of these improvements vary across models. More capable models yield more coherent and balanced coordination effects across states, whereas less capable models exhibit greater variability in policy quality and regional outcomes. Therefore, LLM reasoning capacity remains essential for effective coordinated policymaking. In addition to LLM itself, system-level designs play a critical role in overall performance. The coordination protocol, such as the number of inter-agent communication rounds, directly affects the degree of strategic alignment across regions. Likewise, planning horizons determine the balance between anticipatory coordination and uncertainty accumulation: too short horizons limit forward-looking reasoning, while overly long horizons introduce additional noise. Therefore, effective coordinated policymaking emerges not from any single component alone, but from the interaction between LLM reasoning capability, coordination design, and planning horizon.

Despite these encouraging results, it is important to clarify the scope and limitations of the proposed framework. In practice, LLM agents operate within the bounds of their prompt design and learned representations~\cite{zhao2024expel, sahoo2024systematic}. When domain knowledge is incomplete, contextual signals are poorly specified, or coordination protocols are suboptimal, agent reasoning may fail to capture critical system dynamics, leading to less effective policy recommendations. Beyond agent reasoning, real-world policy effectiveness may be further constrained by political, administrative, and logistical factors, as well as imperfect individual compliance, where travel and mobility decisions do not fully align with prescribed interventions. These execution and compliance frictions introduce discrepancies between simulated policies and their realized outcomes. Moreover, the policy interventions examined in this study operate primarily at a macro level and focus on regulating inter-regional mobility, representing only one dimension of real-world pandemic control. Real-world interventions often involve a broader and more heterogeneous set of measures, such as vaccination campaigns, mask mandates, testing strategies, and their combined deployment. Extending the framework to incorporate diverse interventions, and to reason over multimodal information, remains an important direction for future research.

Beyond pandemic control, the proposed framework may also inform a broader class of coordination problems involving interconnected regions or sectors that share resources, such as regional power grid load balancing~\cite{rahman2013survey}, traffic routing in urban networks~\cite{garrosi2017geo}, or coordination of humanitarian aid flows~\cite{akhtar2012coordination}. In these domains, localized decision-making and inter-regional spillovers similarly give rise to system-level externalities that benefit from coordinated reasoning. By embedding domain-specific simulation models into each agent’s reasoning process and aligning decisions through structured communication, LLM agents could support coordinated decision-making in a principled manner. 
Nevertheless, each new application would require careful modeling of domain dynamics and thoughtful prompt design. In this sense, our study does not claim direct applicability across domains, but rather illustrates a general paradigm: LLM agents can act as synthetic policy advisors that enable systematic reasoning over interconnected decisions. As LLM technology continues to evolve, more capable multi-agent architectures~\cite{guo2024large} may further enhance decision support in complex policy settings by complementing expert judgment and traditional optimization tools.

\newpage
\section*{Methods}\label{sec4}

\subsection*{Raw Data Description}
Multiple categories of data are applied within the LLM multi-agent policymaking framework, including spatial, epidemiological time-series, state-to-state human mobility flow, and public health policy data. All spatial data are provided at the state level, and all time-varying data are provided at daily resolution.

\paragraph{Epidemiological time-series data.}
State-level epidemiological time series were collected from April 12, 2020 to December 31, 2020. 
Each daily record is indexed by the state name and includes the latitude and longitude of the state’s geographic centroid for visualization purposes. 
The dataset reports Confirmed cases and Deaths, which may include both confirmed and probable counts when available. 
Recovered cases are compiled from the COVID Tracking Project and may underestimate the true\cite{dong2020interactive}. 
Active cases are calculated as confirmed cases minus recovered cases minus deaths.

\paragraph{Human mobility flow data.}
Human mobility flows were obtained as daily origin-destination (OD) records covering January 1, 2019, to April 15, 2021. 
Each record describes movements from an origin state to a destination state on a given date. 
For both origin and destination states, the latitude and longitude of the geographic centroid are provided. 
Two flow measures are included: the detected number of visitors traveling from origin to destination, and a corresponding population-flow estimate inferred from the detected flow. The latter is used as the daily state-to-state flow in this study.

\paragraph{Public health policy data.}
Public health policy data were compiled for each U.S. state from March 2020 to 2021 and stored as CSV files with daily updates. 
For each state and date, the dataset records policy categories and corresponding implementation details, including reopening status, mask mandates, health/medical measures, and social distancing, etc. 
Specifically, mobility-related policies are extracted with the assistance of an LLM and used as the ground-truth policy set (see the Supplementary Notes 1.4 and Tables 4-10 for the detailed prompts and corresponding results, respectively).

\subsection*{Framework Design}
We propose a decentralized, LLM multi-agent policymaking framework in which each administrative region is represented by an autonomous policy agent. The framework is designed to support coordinated sequential decision-making under heterogeneous regional conditions and explicit inter-regional coupling induced by population mobility (Fig.~\ref{fig:fig1}c).

\paragraph{LLM agents as policy decision units.} Large language models can be viewed as parametric models that learn a conditional distribution over sequences, $p_{\theta}(y \mid x)$, where $x$ denotes a contextual input expressed in natural language and $y$ denotes the generated response. Through large-scale pretraining on diverse corpora, LLMs acquire latent representations that implicitly encode semantic regularities and causal patterns, enabling contextual reasoning over heterogeneous and partially structured information. In this sense, an LLM can be regarded as a flexible reasoning operator, $y = f_{\theta}(x)$, capable of synthesizing evidence, inferring relationships, and evaluating trade-offs without explicitly defined decision rules.
An \emph{agent} augments such a reasoning model with interaction and persistence. Formally, an agent can be defined as a tuple $\mathrm{A} = (\mathrm{O}, f_{\theta}, \mathrm{P}, \mathrm{T})$, where $\mathrm{O}$ denotes the observation space, $f_{\theta}$ is the LLM-based reasoning function, $\mathrm{P}$ is the policy action space, and $\mathrm{T}$ is a transition operator that maps actions to environment state changes. At each decision epoch $t$, the agent receives an observation $o_t \in \mathrm{O}$, produces an action $u_t = f_{\theta}(o_t)$, and influences system evolution through $\mathrm{T}$. This abstraction enables sequential decision-making with feedback, transforming LLMs from static predictors into goal-directed decision units.

\paragraph{Agent-based policymaking architecture.}
Let $\mathrm{R}=\{1,\dots,N\}$ denote the set of states. Each region $i \in \mathrm{R}$ is associated with an LLM-based policy agent $\mathrm{A}_i$. At each decision epoch $t$, agent $\mathrm{A}_i$ observes a structured state representation
\begin{equation}
\mathrm{O}_i(t)
=
\bigl(
\mathbf{x}_i(t),\;
\{\mathbf{x}_j(t)\}_{j \in \mathrm{N}_i},\;
\mathbf{m}_{\cdot \rightarrow i}(t),\;
\boldsymbol{\theta}_i
\bigr),
\end{equation}
where $\mathbf{x}_i(t)$ summarizes the local pandemic state of region $i$, $\mathrm{N}_i$ denotes neighboring or epidemiologically coupled regions, $\mathbf{m}_{\cdot \rightarrow i}(t)$ represents projected inbound mobility flows from other regions, and $\boldsymbol{\theta}_i$ encodes region-specific attributes such as population structure and healthcare capacity. These heterogeneous inputs are standardized and embedded into structured prompts that serve as the perception interface of each agent. Each agent reasons over its observation using an LLM to generate a policy proposal. Formally, the agent implements a mapping $\pi_i:\; \mathrm{O}_i(t) \;\longrightarrow\; \mathrm{P}_i(t)$, where $\mathrm{P}_i(t)$ denotes the space of admissible policy actions for region $i$. In this work, actions primarily operate on inter-regional mobility as a controllable policy lever, although the framework is not limited to a single intervention modality.

\paragraph{Multi-agent coordination and communication.}
A key feature of the framework is explicit coordination among regional agents. Agents do not act in isolation; instead, they exchange summarized policy-relevant signals, including projected pandemic trends, risk assessments, and proposed interventions. This inter-agent communication induces a coupled decision process in which each agent conditions its policy on the anticipated actions of others. We model coordination as an iterative message-passing process. At decision epoch $t$, each agent $\mathrm{A}_i$ broadcasts a message $\mathrm{M}_i(t) = \phi\bigl(\mathrm{O}_i(t),\; \mathrm{P}_i(t) \bigr)$ to other agents, where $\phi(\cdot)$ extracts salient information such as planned mobility adjustments or inferred transmission risk. Received messages $\{\mathrm{M}_j(t)\}_{j \neq i}$ are incorporated into the subsequent prompt context of agent $i$, allowing the LLM to reason about cross-regional spillovers and align decisions with system-level objectives. This communication mechanism enables decentralized yet coordinated policymaking: agents retain autonomy over regional decisions while implicitly negotiating policies through shared information and iterative reasoning.

\paragraph{Policy parameterization and feasibility constraints.}
To ensure realism and comparability with observed policies, agent actions are parameterized under explicit feasibility constraints. For the primary intervention studied—\emph{Temporal Inflow Reallocation (TIR)}—each agent outputs a normalized allocation vector
\begin{equation}
\boldsymbol{p}_i
=
(p_{i,1},\dots,p_{i,H}),
\qquad
\sum_{h=1}^{H} p_{i,h} = 1,
\quad
p_{i,h} > 0,
\end{equation}
which specifies how the total inbound mobility volume over a planning horizon of $H$ weeks is redistributed across time. Importantly, the aggregate inflow remains fixed, preserving economic and social plausibility while allowing agents to modulate \emph{when} mobility occurs. The same agent architecture supports alternative intervention classes, such as \textit{Spatial Inflow Suppression} or \textit{Targeted Inbound Screening}, by modifying the action space $\mathrm{P}_i(t)$ while keeping the perception, reasoning, and coordination modules unchanged. This modularity enables systematic comparison across policy mechanisms within a unified framework.

\paragraph{Closed-loop interaction with the simulation environment.}
The proposed framework operates through an explicit closed-loop interaction between policy agents and a mechanistic pandemic simulation environment. At each decision epoch $t$, the simulator first evolves the pandemic dynamics forward using a calibrated pandemic transmission model,
\begin{equation}
\mathbf{X}(t+1) = \mathrm{T}\bigl(\mathbf{X}(t), \mathbf{M}(t)\bigr),
\end{equation}
where $\mathrm{T}$ denotes the transition operator induced by the mechanistic pandemic simulator; $\mathbf{X}(t)=\{\mathbf{x}_i(t)\}_{i\in\mathrm{R}}$ denotes the collection of region-level SEIQRD states and $\mathbf{M}(t)=\{\mathbf{m}_{\cdot \rightarrow i}(t)\}_{i\in\mathrm{R}}$ collects region-level inbound mobility flows. From the simulated SEIQRD trajectories, evaluation metrics such as active cases, incidence rate, and mortality are derived and provided to the agents as part of their observations. Conditioned on these evaluated pandemic indicators and the current mobility flows, LLM agents engage in multi-agent communication and reasoning to determine coordinated policy actions. For region $i$, the agent outputs a parameterized intervention $\boldsymbol{p}_i(t) \in \mathrm{P}_i(t)$ that reallocates inbound mobility over the planning horizon. Let $M_i$ denote the total inbound mobility volume of region $i$, which is preserved under a total volume constraint. The realized mobility after policy enactment is given by
\begin{equation}
m_{i,h}(t) = p_{i,h}(t)\cdot M_i,
\qquad
\sum_{h=1}^{H} m_{i,h}(t) = M_i .
\end{equation}
Once enacted, the updated mobility configuration $\mathbf{M}(t+1)$ is fed back into the pandemic simulator, which re-evaluates disease transmission under the new policy setting. The resulting pandemic states $\mathbf{X}(t+1)$ form the input to the next decision epoch, completing a closed observation--decision--transition loop. This iterative interaction enables agents to anticipate downstream consequences of policy choices, continuously adapt to evolving pandemic conditions, and refine coordinated interventions over time.

Overall, the proposed framework integrates structured data inputs, LLM agent reasoning, inter-agent communication, and constrained policy execution into a unified multi-agent decision system. This design allows coordinated, forward-looking policymaking to emerge endogenously from decentralized agent interactions, providing a flexible foundation for AI-assisted pandemic control.

\subsection*{Epidemiological Modeling and Simulation}
State-level pandemic dynamics are simulated using a discrete-time SEIQRD model with daily updates.
For each state $i$, the population is partitioned into six compartments: susceptible $S_i$, exposed $E_i$, infected $I_i$, quarantined $Q_i$, recovered $R_i$, and deceased $D_i$.
The model captures: 1) latent infection prior to infectiousness ($E$), 2) isolation of detected infected individuals ($Q$), and 3) mortality ($D$) \cite{tan2022statistical,balcan2009multiscale}. 
%

First, $t$ denotes the day index, and $\Delta t=1$day. The within-state force of infection is
$\lambda_{i,t} = \beta_{i}^I\frac{I_{i,t}}{N_{i,t}} + \beta_{i}^Q \frac{Q_{i,t}}{N_{i,t}}$,
where $N_{i,t} = S_{i,t} + E_{i,t}+I_{i,t}+Q_{i,t}+R_{i,t}$ is the living population, and the $\beta_{i}^I$ and $\beta_{i}^Q$ is the transmission rate for infectious and quarantined individuals, respectively. Quarantined individuals are assumed to have lower onward transmission, i.e., $\beta_{i}^Q <\beta_{i}^I$. 

Next, to incorporate inter-state mobility
, $m_{j \to i,t}$ denotes the inbound flow from origin state $j$ to destination state $i$ \cite{wang2018inferring}. 
Inter-state mobility affects the epidemic dynamics by redistributing individuals across compartments.
The resulting daily compartment updates are:
\begin{align}
S_{i,t+1} &= S_{i,t}
- \left( \lambda_{i,t} S_{i,t} \right) + \sum_{j \neq i} m_{j \to i,t}\frac{S_{j,t}}{N_{j,t}} - \sum_{j \neq i} m_{i \to j,t}\frac{S_{i,t}}{N_{i,t}} , \\
E_{i,t+1} &= E_{i,t}
+ \left( \lambda_{i,t} S_{i,t} \right)
- \sigma_i E_{i,t} + \sum_{j \neq i} m_{j \to i,t}\frac{E_{j,t}}{N_{j,t}} - \sum_{j \neq i} m_{i \to j,t}\frac{E_{i,t}}{N_{i,t}}, \\
I_{i,t+1} &= I_{i,t}
+ \sigma_i E_{i,t}
- \left( \delta_{i,t} + \gamma_{i,t} + \mu_{i,t} \right) I_{i,t} + \sum_{j \neq i} m_{j \to i,t}\frac{I_{j,t}}{N_{j,t}} - \sum_{j \neq i} m_{i \to j,t}\frac{I_{i,t}}{N_{i,t}}, \\
Q_{i,t+1} &= Q_{i,t}
+ \delta_{i,t} I_{i,t}
- \left( \gamma_{i,t} + \mu_{i,t} \right) Q_{i,t}, \\
R_{i,t+1} &= R_{i,t}
+ \gamma_{i,t} I_{i,t}
+ \gamma_{i,t} Q_{i,t} + \sum_{j \neq i} m_{j \to i,t}\frac{R_{j,t}}{N_{j,t}} - \sum_{j \neq i} m_{i \to j,t}\frac{R_{i,t}}{N_{i,t}}, \\
D_{i,t+1} &= D_{i,t}
+ \mu_{i,t} I_{i,t}
+ \mu_{i,t} Q_{i,t}.
\end{align}
Here, $\sigma_i$ denotes the progression rate from the exposed to the infectious compartment,
$\delta_{i,t}$ represents the detection rate,
$\gamma_{i,t}$ is the recovery rate for infectious and quarantined individuals, 
and $\mu_{i,t}$ denotes the mortality rate. Quarantined individuals are assumed not to be allowed to move across states.
All parameters are calibrated using the ground-truth simulation (see Supplementary Note 1.5 for the detailed calibration strategy).

Policy interventions affect the pandemic dynamics by modulating
1) the inter-state mobility matrix $m_{j \to i,t}$,
and 2) the effective detection rate $\delta_{i,t}$ through targeted screening,
thereby altering the transition from $E$ and $I$ to $Q$.

\subsection*{Evaluation Metrics in Pandemic Control}
\paragraph{Cumulative metrics.} Total infections are measured using the cumulative quarantined count at the end of the simulation, reflecting the same case-recording practice adopted in official statistics to ensure fair comparison. Total deaths are defined as the cumulative number of deaths at the end of the simulation.

\paragraph{Temporal indicators.} Daily incidence rate (IR), death rate (DR), and active case rate (ACR) are used to evaluate the performance of intervention in pandemic control \cite{cabezudo2020incidence}. IR  measures the number of newly confirmed infections per day, capturing the short-term transmission intensity of the pandemic \cite{xiong2020mobile}.
\begin{equation}
\mathrm{IR}_{i,t}
= \frac{Q_{i,t+1}-Q_{i,t}}{N_i}.
\end{equation}
ACR quantifies the number of currently active confirmed infections, defined as the population remaining in quarantined compartments. ACR reflects the ongoing disease burden on the healthcare system and is closely associated with hospital occupancy and resource utilization \cite{li2020active}.
\begin{equation}
\mathrm{ACR}_{i,t}
= \frac{Q_{i,t}}{N_i}.
\end{equation}
DR measures daily new deaths per capita and serves as a key indicator of severe outcomes
\begin{equation}
\mathrm{DR}_{i,t}
= \frac{D_{i,t+1} - D_{i,t}}{N_i},
\end{equation}
The effectiveness of each policy is also evaluated using the average IR, ACR, and DR across all time periods for a overall assessment.

\subsection*{Estimation of the Effective Reproduction Number $R_t$}
The effective (time-varying) reproduction number $R_t$ is estimated from the daily observed incidence series using a renewal-process formulation with Bayesian updating, following the widely used framework for real-time transmissibility inference \cite{cori2013new,thompson2019improved}. $Q'_t=Q_t-Q_{t-1}$ denotes the number of newly infected (incident) cases on day $t$. Incident infections on day $t$ arise from past infections weighted by a discretized serial interval distribution.

\paragraph{Discrete serial interval.}
The serial interval is modeled as a Gamma distribution with mean $\mu_{\mathrm{SI}}=5$ days and standard deviation $\sigma_{\mathrm{SI}}=2$ days, reflecting the typical delay between symptom onset in a exposed case and symptom onset in infected cases\cite{yin2021parameter}. The corresponding Gamma shape and scale parameters are
$
k = \left(\frac{\mu_{\mathrm{SI}}}{\sigma_{\mathrm{SI}}}\right)^2$
,$\theta = \frac{\sigma_{\mathrm{SI}}^2}{\mu_{\mathrm{SI}}}.
$
This distribution is discretized into daily probabilities over lags $s=1,\dots,S$ with $S=20$ days:
\begin{equation}
w_s \;=\; F_{\Gamma}(s; k,\theta) - F_{\Gamma}(s-1; k,\theta),
\qquad s=1,\dots,S,
\end{equation}
where $F_{\Gamma}(\cdot;k,\theta)$ is the Gamma cumulative distribution function. The weights are normalized such that $\sum_{s=1}^{S} w_s = 1$.

\paragraph{Renewal equation and infectiousness.}
Define the total infectiousness (renewal intensity) on day $u$ as
$
\Lambda_u \;=\; \sum_{s=1}^{S} Q'_{u-s}\, w_s,
$
with the convention $Q'_{u-s}=0$ when $u-s<0$. Under the renewal model, conditional on $R_u$, the incidence satisfies
\begin{equation}
Q'_u \mid R_u \;\sim\; \mathrm{Poisson}(R_u\,\Lambda_u).
\end{equation}

\paragraph{Sliding-window Bayesian inference.}
To stabilize estimation, $R_t$ is assumed to be constant within a sliding window of length $W=21$ days. For each evaluation time $t\ge W$, 
$
Q'^{(t)}_{\Sigma} \;=\; \sum_{u=t-W}^{t-1} Q'_u,
$, and 
$\Lambda^{(t)}_{\Sigma} \;=\; \sum_{u=t-W}^{t-1} \Lambda_u .$
A Gamma prior is placed on $R_t$,
$R_t \sim \mathrm{Gamma}(a,b),$
where $a=1$ and $b=1$ denote the prior shape and rate parameters, respectively. Combining the Poisson likelihood with the Gamma prior yields a conjugate Gamma posterior\cite{ghahramani2015probabilistic}:
\begin{equation}
R_t \mid \{Q'_u\}_{u=t-W}^{t-1}
\;\sim\;
\mathrm{Gamma}\!\left(a + Q'^{(t)}_{\Sigma},\; b + \max\!\big(\Lambda^{(t)}_{\Sigma},\varepsilon\big)\right),
\label{eq:rt_posterior}
\end{equation}
where $\varepsilon$ is a small positive constant to avoid numerical instability.

\paragraph{Point and interval estimates.}
The posterior mean of $R_t$ is
$
\widehat{R}_t \;=\; \mathbb{E}[R_t \mid \cdot]
\;=\;
\frac{a + Q'^{(t)}_{\Sigma}}{b + \max(\Lambda^{(t)}_{\Sigma},\varepsilon)}.
$
A $95\%$ credible interval $\big[R_t^{\mathrm{low}}, R_t^{\mathrm{high}}\big]$ is kept, using the $2.5\%$ and $97.5\%$ posterior quantiles of the Gamma distribution in Eq.~\eqref{eq:rt_posterior}. The above procedure produces a time series of $\widehat{R}_t$ and its uncertainty bounds.

\subsection*{Policy Type Classification}
To characterize the qualitative patterns of adaptive policy responses, policy outputs are categorized into three representative types based on their temporal allocation under TIR \cite{nouvellet2021reduction}. 
Taking 6-week TIR for illustration, each policy response specifies weekly reallocation levels for each state over a six-week planning horizon. 

First, $\mathbf{p}_i = [p_{i,1},\dots,p_{i,6}]$ denotes the six-week policy sequence for state $i$, where $p_{i,h}$ indicates the proportion of inbound mobility allocated to week $h$ within the planning horizon
and $\sum_{h\in\{1,...,6\}}{p_{i,h}=1}$. 
The six-length proportions are aggregated into three consecutive periods, corresponding to the early, middle, and late phases of the planning horizon.
$[\bar{p}_{i}^{1},\ \bar{p}_{i}^{2},\ \bar{p}_{i}^{3}]
$
represent weeks 1–2, 3–4, and 5–6, respectively. Here, the subscript $t$ is omitted for clarity.

Next, each policy instance is classified into one of three types:
\begin{itemize}
\item \textbf{Strict-first}: policies allowing relatively low inbound flow in the early period but higher inflow in the late period, defined by
$
\bar{p}_{i}^{1} \leq 0.3 \text{ and }  \bar{p}_{i}^{3} \geq 0.4;
$
\item \textbf{Relaxed-first}: policies characterized by relaxed early interventions followed by restriction in later stages, defined by
$
\bar{p}_{i}^{1} \geq 0.4 \text{ and } \bar{p}_{i}^{3} \leq 0.3;
$
\item \textbf{Balanced}: all remaining policies that do not exhibit a pronounced shift toward either early strictness or late strictness.
\end{itemize}

\subsection*{Feature Attribution on Policy Type}

To interpret the determinants underlying the emergence of \emph{strict-first} policy patterns, a feature attribution analysis is conducted using Shapley values. The propensity of adopting a strict-first policy is modeled as a supervised learning task, in which a tree-based ensemble model maps contextual features to the corresponding policy classification outcome \cite{chan2008evaluation}.

Feature contributions are quantified using Shapley additive explanations, which attribute the prediction of a model to individual input features based on cooperative game theory\cite{zhao2025safetraffic}. $f(\mathbf{x})$ denotes the trained prediction function and $\mathrm{F}=\{1,\dots,M\}$ the set of input features. The Shapley value for feature $j$ is defined as
\begin{equation}
\phi_j
=
\sum_{S \subseteq \mathrm{F} \setminus \{j\}}
\frac{|S|!\,(M-|S|-1)!}{M!}
\left[
f(\mathbf{x}_{S \cup \{j\}}) - f(\mathbf{x}_{S})
\right],
\end{equation}
where $S$ denotes a subset of features excluding $j$, and $\mathbf{x}_{S}$ represents the input vector in which only features in $S$ are present. This formulation measures the expected marginal contribution of feature $j$ across all possible feature coalitions.

\subsection*{Equity Coefficient Calculation}
To quantify the equity of policy-induced benefits across states, an \emph{equity coefficient} is defined based on the Gini index, which measures the dispersion of improvements among states \cite{scheffer2017inequality}. The coefficient is computed separately for total infections and total deaths.
\paragraph{Improvement of state.}
$G^{\mathrm{inf}}_i$ and $G^{\mathrm{dea}}_i$ is the cumulative total infections and deaths observed in the ground-truth scenario ($i \in \{1,\dots,N\}$ index states), while  $A^{\mathrm{inf}}_i$ and $A^{\mathrm{dea}}_i$ is the corresponding cumulative outcomes under the agent-based policy at the end of the simulation. The relative improvement in total infections and deaths for state $i$ is
\begin{equation}
\Delta^{\mathrm{inf}}_i
= \frac{G^{\mathrm{inf}}_i - A^{\mathrm{inf}}_i}
{\max\!\left(|G^{\mathrm{inf}}_i|,\varepsilon\right)},
\qquad
\Delta^{\mathrm{dea}}_i
= \frac{G^{\mathrm{dea}}_i - A^{\mathrm{dea}}_i}
{\max\!\left(|G^{\mathrm{dea}}_i|,\varepsilon\right)},
\end{equation}
where $\varepsilon$ is a small positive constant to avoid numerical instability. To ensure that the equity measure reflects, only the distribution of benefits, we focus on non-negative improvements:
\begin{equation}
\Delta^{\mathrm{inf},+}_i = \max(\Delta^{\mathrm{tot}}_i, 0),
\qquad
\Delta^{\mathrm{dea},+}_i = \max(\Delta^{\mathrm{dea}}_i, 0).
\end{equation}

\paragraph{Gini-based equity coefficient.}
Given a non-negative improvement vector $\mathbf{x} = (x_1,\dots,x_N)$, the Gini coefficient is defined as
\begin{equation}
G(\mathbf{x})
= \frac{1}{N \sum_{r=1}^{N} x_r}
\sum_{r=1}^{N} (2r - N - 1)\, x_{r},
\end{equation}
where $x_{r}$ denotes the $r$-th smallest element of $\mathbf{x}$. A smaller Gini coefficient indicates a more equal distribution of improvements across states \cite{blesch2022measuring}.

The equity coefficient is defined as
$
\mathrm{E}(\mathbf{x}) = 1 - G(\mathbf{x}),
$
such that larger values indicate greater equity, with $\mathrm{E}=1$ corresponding to perfectly equal benefit allocation. Applying this definition to infection and death reductions yields
\begin{equation}
\mathrm{E}^{\mathrm{inf}} = 1 - G\!\left(\{\Delta^{\mathrm{inf},+}_i\}_{i=1}^{N}\right),
\qquad
\mathrm{E}^{\mathrm{dea}} = 1 - G\!\left(\{\Delta^{\mathrm{dea},+}_i\}_{i=1}^{N}\right).
\end{equation}

\subsection*{Forecasting Subsequent 180-Day Cumulative Infections and Deaths}
To facilitate a consistent comparison of long-term outcomes across policies, cumulative infections and deaths are extrapolated up to 180 days beyond the simulation horizon using a local trend-based approach \cite{xia2023future, althoff2025countrywide}. $C_t$ denotes the cumulative number of infections or deaths at day $t$. The local daily growth rate is estimated as the average increment over the most recent 14 days,
\begin{equation}
\bar{c}_t = \frac{1}{14}\sum_{i=1}^{14} \left(C_{t-i+1} - C_{t-i}\right).
\end{equation}
Cumulative outcomes are extrapolated recursively for $h = 1,\dots,180$ days according to
\begin{equation}
C_{t+h} = C_{t+h-1} + \bar{c}_{t}, h \in \{1,\dots,180\}.
\end{equation}

\section*{Competing interests} 
The authors declare no competing interests.

\newpage
\bibliography{sample}

@article{asthana2025decision,
  title={Decision-making under epistemic, strategic and institutional uncertainty during COVID-19: findings from a six-country empirical study},
  author={Asthana, Sumegha and Mukherjee, Sanjana and Phelan, Alexandra L and Gobir, Ibrahim B and Woo, JJ and Wenham, Clare and Husain, Mohammad Mushtuq and Shirin, Tahmina and Govender, Nevashan and Al Nsour, Mohannad and others},
  journal={BMJ Global Health},
  volume={10},
  number={2},
  year={2025},
  publisher={BMJ Publishing Group Ltd}
}

@article{ferguson2006strategies,
  title={Strategies for mitigating an influenza pandemic},
  author={Ferguson, Neil M and Cummings, Derek AT and Fraser, Christophe and Cajka, James C and Cooley, Philip C and Burke, Donald S},
  journal={Nature},
  volume={442},
  number={7101},
  pages={448--452},
  year={2006},
  publisher={Nature Publishing Group UK London}
}

@article{holtz2020interdependence,
  title={Interdependence and the cost of uncoordinated responses to COVID-19},
  author={Holtz, David and Zhao, Michael and Benzell, Seth G and Cao, Cathy Y and Rahimian, Mohammad Amin and Yang, Jeremy and Allen, Jennifer and Collis, Avinash and Moehring, Alex and Sowrirajan, Tara and others},
  journal={Proceedings of the National Academy of Sciences},
  volume={117},
  number={33},
  pages={19837--19843},
  year={2020},
  publisher={National Academy of Sciences}
}

@article{domenici2022fragmented,
  title={The fragmented nature of pandemic decision-making: A comparative and multilevel legal analysis},
  author={Domenici, Irene and Villarreal, Pedro A},
  journal={European Journal of Health Law},
  volume={29},
  number={1},
  pages={1--5},
  year={2022},
  publisher={Brill Nijhoff}
}

@article{sachs2022lancet,
  title={The Lancet Commission on lessons for the future from the COVID-19 pandemic},
  author={Sachs, Jeffrey D and Karim, Salim S Abdool and Aknin, Lara and Allen, Joseph and Brosb{\o}l, Kirsten and Colombo, Francesca and Barron, Gabriela Cuevas and Espinosa, Mar{\'\i}a Fernanda and Gaspar, Vitor and Gaviria, Alejandro and others},
  journal={The Lancet},
  volume={400},
  number={10359},
  pages={1224--1280},
  year={2022},
  publisher={Elsevier}
}

@article{graff2020spread,
  title={The spread of COVID-19 shows the importance of policy coordination},
  author={Graff Zivin, Joshua and Sanders, Nicholas},
  journal={Proceedings of the National Academy of Sciences},
  volume={117},
  number={52},
  pages={32842--32844},
  year={2020},
  publisher={National Academy of Sciences}
}

@article{flaxman2020estimating,
  title={Estimating the effects of non-pharmaceutical interventions on COVID-19 in Europe},
  author={Flaxman, Seth and Mishra, Swapnil and Gandy, Axel and Unwin, H Juliette T and Mellan, Thomas A and Coupland, Helen and Whittaker, Charles and Zhu, Harrison and Berah, Tresnia and Eaton, Jeffrey W and others},
  journal={Nature},
  volume={584},
  number={7820},
  pages={257--261},
  year={2020},
  publisher={Nature Publishing Group UK London}
}

@article{chang2020modelling,
  title={Modelling transmission and control of the COVID-19 pandemic in Australia},
  author={Chang, Sheryl L and Harding, Nathan and Zachreson, Cameron and Cliff, Oliver M and Prokopenko, Mikhail},
  journal={Nature communications},
  volume={11},
  number={1},
  pages={5710},
  year={2020},
  publisher={Nature Publishing Group UK London}
}

@article{mckee2025power,
  title={The power of artificial intelligence for managing pandemics: A primer for public health professionals},
  author={McKee, Martin and Rosenbacke, Rikard and Stuckler, David},
  journal={The International journal of health planning and management},
  volume={40},
  number={1},
  pages={257--270},
  year={2025},
  publisher={Wiley Online Library}
}

@article{bicher2022supporting,
  title={Supporting COVID-19 policy-making with a predictive epidemiological multi-model warning system},
  author={Bicher, Martin and Zuba, Martin and Rainer, Lukas and Bachner, Florian and Rippinger, Claire and Ostermann, Herwig and Popper, Nikolas and Thurner, Stefan and Klimek, Peter},
  journal={Communications medicine},
  volume={2},
  number={1},
  pages={157},
  year={2022},
  publisher={Nature Publishing Group UK London}
}

@article{sacco2023proactive,
  title={Proactive vs. reactive country responses to the COVID-19 pandemic shock},
  author={Sacco, Pier Luigi and Valle, Francesco and De Domenico, Manlio},
  journal={PLOS Global Public Health},
  volume={3},
  number={1},
  pages={e0001345},
  year={2023},
  publisher={Public Library of Science San Francisco, CA USA}
}

@article{prem2020effect,
  title={The effect of control strategies to reduce social mixing on outcomes of the COVID-19 epidemic in Wuhan, China: a modelling study},
  author={Prem, Kiesha and Liu, Yang and Russell, Timothy W and Kucharski, Adam J and Eggo, Rosalind M and Davies, Nicholas and Flasche, Stefan and Clifford, Samuel and Pearson, Carl AB and Munday, James D and others},
  journal={The lancet public health},
  volume={5},
  number={5},
  pages={e261--e270},
  year={2020},
  publisher={Elsevier}
}

@article{kissler2020projecting,
  title={Projecting the transmission dynamics of SARS-CoV-2 through the postpandemic period},
  author={Kissler, Stephen M and Tedijanto, Christine and Goldstein, Edward and Grad, Yonatan H and Lipsitch, Marc},
  journal={Science},
  volume={368},
  number={6493},
  pages={860--868},
  year={2020},
  publisher={American Association for the Advancement of Science}
}

@article{pacces2020diversity,
  title={From diversity to coordination: A European approach to COVID-19},
  author={Pacces, Alessio M and Weimer, Maria},
  journal={European journal of risk regulation},
  volume={11},
  number={2},
  pages={283--296},
  year={2020},
  publisher={Cambridge University Press}
}

@article{ruktanonchai2020assessing,
  title={Assessing the impact of coordinated COVID-19 exit strategies across Europe},
  author={Ruktanonchai, Nick Warren and Floyd, Jessica R and Lai, Shengjie and Ruktanonchai, Corrine Warren and Sadilek, Adam and Rente-Lourenco, Pedro and Ben, Xue and Carioli, Alessandra and Gwinn, Joshua and Steele, Jessica E and others},
  journal={Science},
  volume={369},
  number={6510},
  pages={1465--1470},
  year={2020},
  publisher={American Association for the Advancement of Science}
}

@article{zhao2023survey,
  title={A survey of large language models},
  author={Zhao, Wayne Xin and Zhou, Kun and Li, Junyi and Tang, Tianyi and Wang, Xiaolei and Hou, Yupeng and Min, Yingqian and Zhang, Beichen and Zhang, Junjie and Dong, Zican and others},
  journal={arXiv preprint arXiv:2303.18223},
  volume={1},
  number={2},
  year={2023}
}

@article{guo2024large,
  title={Large language model based multi-agents: A survey of progress and challenges},
  author={Guo, Taicheng and Chen, Xiuying and Wang, Yaqi and Chang, Ruidi and Pei, Shichao and Chawla, Nitesh V and Wiest, Olaf and Zhang, Xiangliang},
  journal={arXiv preprint arXiv:2402.01680},
  year={2024}
}

@article{zhao2025safetraffic,
  title={SafeTraffic Copilot: adapting large language models for trustworthy traffic safety assessments and decision interventions},
  author={Zhao, Yang and Wang, Pu and Zhao, Yibo and Du, Hongru and Yang, Hao Frank},
  journal={Nature Communications},
  volume={16},
  number={1},
  pages={8846},
  year={2025},
  publisher={Nature Publishing Group UK London}
}

@article{du2025advancing,
  title={Advancing real-time infectious disease forecasting using large language models},
  author={Du, Hongru and Zhao, Yang and Zhao, Jianan and Xu, Shaochong and Lin, Xihong and Chen, Yiran and Gardner, Lauren M and Yang, Hao ‘Frank’},
  journal={Nature Computational Science},
  pages={1--14},
  year={2025},
  publisher={Nature Publishing Group US New York}
}

@article{kwok2024utilizing,
  title={Utilizing large language models in infectious disease transmission modelling for public health preparedness},
  author={Kwok, Kin On and Huynh, Tom and Wei, Wan In and Wong, Samuel YS and Riley, Steven and Tang, Arthur},
  journal={Computational and Structural Biotechnology Journal},
  volume={23},
  pages={3254--3257},
  year={2024},
  publisher={Elsevier}
}

@article{williams2023epidemic,
  title={Epidemic modeling with generative agents},
  author={Williams, Ross and Hosseinichimeh, Niyousha and Majumdar, Aritra and Ghaffarzadegan, Navid},
  journal={arXiv preprint arXiv:2307.04986},
  year={2023}
}

@inproceedings{zhou2023traditional,
  title={Traditional Chinese medicine epidemic prevention and treatment question-answering model based on LLMs},
  author={Zhou, Zongzhen and Yang, Tao and Hu, Kongfa},
  booktitle={2023 IEEE International Conference on Bioinformatics and Biomedicine (BIBM)},
  pages={4755--4760},
  year={2023},
  organization={IEEE}
}

@inproceedings{consoli2024epidemic,
  title={Epidemic Information Extraction for Event-Based Surveillance Using Large Language Models},
  author={Consoli, Sergio and Markov, Peter and Stilianakis, Nikolaos I and Bertolini, Lorenzo and Gallardo, Antonio Puertas and Ceresa, Mario},
  booktitle={International Congress on Information and Communication Technology},
  pages={241--252},
  year={2024},
  organization={Springer Nature Singapore Singapore}
}

@inproceedings{saeed2024llm4cast,
  title={LLM4cast: repurposed LLM for viral disease forecasting},
  author={Saeed, Farah and Aldosari, Mohammed and Arpinar, Ismailcem Budak and Miller, John A},
  booktitle={2024 IEEE International Conference on Big Data (BigData)},
  pages={1428--1433},
  year={2024},
  organization={IEEE}
}

@article{chopra2024limits,
  title={On the limits of agency in agent-based models},
  author={Chopra, Ayush and Kumar, Shashank and Giray-Kuru, Nurullah and Raskar, Ramesh and Quera-Bofarull, Arnau},
  journal={arXiv preprint arXiv:2409.10568},
  year={2024}
}

@article{perofsky2024impacts,
  title={Impacts of human mobility on the citywide transmission dynamics of 18 respiratory viruses in pre-and post-COVID-19 pandemic years},
  author={Perofsky, Amanda C and Hansen, Chelsea L and Burstein, Roy and Boyle, Shanda and Prentice, Robin and Marshall, Cooper and Reinhart, David and Capodanno, Ben and Truong, Melissa and Schwabe-Fry, Kristen and others},
  journal={Nature communications},
  volume={15},
  number={1},
  pages={4164},
  year={2024},
  publisher={Nature Publishing Group UK London}
}

@article{lu2023understanding,
  title={Understanding mobility change in response to COVID-19: A Los Angeles case study},
  author={Lu, Yougeng and Giuliano, Genevieve},
  journal={Travel Behaviour and Society},
  volume={31},
  pages={189--201},
  year={2023},
  publisher={Elsevier}
}

@article{van2025four,
  title={Four million COVID-19 cases in four European cross-border regions reveal minimal cross-border transmission effect on domestic burden: space-time cluster analysis},
  author={van der Zanden, Brigitte and Kauhl, Boris and Hackert, Volker and Hoebe, Christian JPA},
  journal={BMC Infectious Diseases},
  volume={25},
  number={1},
  pages={1528},
  year={2025},
  publisher={Springer}
}

@article{althouse2023unintended,
  title={The unintended consequences of inconsistent closure policies and mobility restrictions during epidemics},
  author={Althouse, Benjamin M and Wallace, Brendan and Case, BKM and Scarpino, Samuel V and Allard, Antoine and Berdahl, Andrew M and White, Easton R and H{\'e}bert-Dufresne, Laurent},
  journal={BMC Global and Public Health},
  volume={1},
  number={1},
  pages={28},
  year={2023},
  publisher={Springer}
}

@article{cori2013new,
  title={A new framework and software to estimate time-varying reproduction numbers during epidemics},
  author={Cori, Anne and Ferguson, Neil M and Fraser, Christophe and Cauchemez, Simon},
  journal={American journal of epidemiology},
  volume={178},
  number={9},
  pages={1505--1512},
  year={2013},
  publisher={Oxford University Press}
}

@article{thompson2019improved,
  title={Improved inference of time-varying reproduction numbers during infectious disease outbreaks},
  author={Thompson, Robin N and Stockwin, Jake E and van Gaalen, Rolina D and Polonsky, Jonny A and Kamvar, Zhian N and Demarsh, P Alex and Dahlqwist, Elisabeth and Li, Siyang and Miguel, Eve and Jombart, Thibaut and others},
  journal={Epidemics},
  volume={29},
  pages={100356},
  year={2019},
  publisher={Elsevier}
}

@article{chan2008evaluation,
  title={Evaluation of Random Forest and Adaboost tree-based ensemble classification and spectral band selection for ecotope mapping using airborne hyperspectral imagery},
  author={Chan, Jonathan Cheung-Wai and Paelinckx, Desir{\'e}},
  journal={Remote Sensing of Environment},
  volume={112},
  number={6},
  pages={2999--3011},
  year={2008},
  publisher={Elsevier}
}

@article{blesch2022measuring,
  title={Measuring inequality beyond the Gini coefficient may clarify conflicting findings},
  author={Blesch, Kristin and Hauser, Oliver P and Jachimowicz, Jon M},
  journal={Nature human behaviour},
  volume={6},
  number={11},
  pages={1525--1536},
  year={2022},
  publisher={Nature Publishing Group UK London}
}

@inproceedings{wang2018inferring,
  title={Inferring metapopulation propagation network for intra-city epidemic control and prevention},
  author={Wang, Jingyuan and Wang, Xiaojian and Wu, Junjie},
  booktitle={Proceedings of the 24th ACM SIGKDD international conference on knowledge discovery \& data mining},
  pages={830--838},
  year={2018}
}

@article{balcan2009multiscale,
  title={Multiscale mobility networks and the spatial spreading of infectious diseases},
  author={Balcan, Duygu and Colizza, Vittoria and Gon{\c{c}}alves, Bruno and Hu, Hao and Ramasco, Jos{\'e} J and Vespignani, Alessandro},
  journal={Proceedings of the national academy of sciences},
  volume={106},
  number={51},
  pages={21484--21489},
  year={2009},
  publisher={National Academy of Sciences}
}

@article{tan2022statistical,
  title={Statistical inference using GLEaM model with spatial heterogeneity and correlation between regions},
  author={Tan, Yixuan and Zhang, Yuan and Cheng, Xiuyuan and Zhou, Xiao-Hua},
  journal={Scientific Reports},
  volume={12},
  number={1},
  pages={16630},
  year={2022},
  publisher={Nature Publishing Group UK London}
}

@article{li2020active,
  title={Active case finding with case management: the key to tackling the COVID-19 pandemic},
  author={Li, Zhongjie and Chen, Qiulan and Feng, Luzhao and Rodewald, Lance and Xia, Yinyin and Yu, Hailiang and Zhang, Ruochen and An, Zhijie and Yin, Wenwu and Chen, Wei and others},
  journal={The lancet},
  volume={396},
  number={10243},
  pages={63--70},
  year={2020},
  publisher={Elsevier}
}

@article{cabezudo2020incidence,
  title={Incidence and case fatality rate of COVID-19 in patients with active epilepsy},
  author={Cabezudo-Garc{\'\i}a, Pablo and Ciano-Petersen, Nicol{\'a}s Lundahl and Mena-V{\'a}zquez, Natalia and Pons-Pons, Gracia and Castro-S{\'a}nchez, Mar{\'\i}a Victoria and Serrano-Castro, Pedro J},
  journal={Neurology},
  volume={95},
  number={10},
  pages={e1417--e1425},
  year={2020},
  publisher={Lippincott Williams \& Wilkins Hagerstown, MD}
}

@article{scheffer2017inequality,
  title={Inequality in nature and society},
  author={Scheffer, Marten and Van Bavel, Bas and van de Leemput, Ingrid A and van Nes, Egbert H},
  journal={Proceedings of the National Academy of Sciences},
  volume={114},
  number={50},
  pages={13154--13157},
  year={2017},
  publisher={National Academy of Sciences}
}

@article{xiong2020mobile,
  title={Mobile device data reveal the dynamics in a positive relationship between human mobility and COVID-19 infections},
  author={Xiong, Chenfeng and Hu, Songhua and Yang, Mofeng and Luo, Weiyu and Zhang, Lei},
  journal={Proceedings of the National Academy of Sciences},
  volume={117},
  number={44},
  pages={27087--27089},
  year={2020},
  publisher={National Academy of Sciences}
}

@article{nouvellet2021reduction,
  title={Reduction in mobility and COVID-19 transmission},
  author={Nouvellet, Pierre and Bhatia, Sangeeta and Cori, Anne and Ainslie, Kylie EC and Baguelin, Marc and Bhatt, Samir and Boonyasiri, Adhiratha and Brazeau, Nicholas F and Cattarino, Lorenzo and Cooper, Laura V and others},
  journal={Nature communications},
  volume={12},
  number={1},
  pages={1090},
  year={2021},
  publisher={Nature Publishing Group UK London}
}

@article{yin2021parameter,
  title={Parameter estimation of the incubation period of COVID-19 based on the doubly interval-censored data model},
  author={Yin, Ming-Ze and Zhu, Qing-Wen and L{\"u}, Xing},
  journal={Nonlinear Dynamics},
  volume={106},
  number={2},
  pages={1347--1358},
  year={2021},
  publisher={Springer}
}

@article{ghahramani2015probabilistic,
  title={Probabilistic machine learning and artificial intelligence},
  author={Ghahramani, Zoubin},
  journal={Nature},
  volume={521},
  number={7553},
  pages={452--459},
  year={2015},
  publisher={Nature Publishing Group UK London}
}

@article{xia2023future,
  title={Future reductions of China’s transport emissions impacted by changing driving behaviour},
  author={Xia, Yingji and Liao, Chenlei and Chen, Xiqun and Zhu, Zheng and Chen, Xiaorui and Wang, Lixing and Jiang, Rui and Stettler, Marc EJ and Angeloudis, Panagiotis and Gao, Ziyou},
  journal={Nature Sustainability},
  volume={6},
  number={10},
  pages={1228--1236},
  year={2023},
  publisher={Nature Publishing Group UK London}
}

@article{althoff2025countrywide,
  title={Countrywide natural experiment links built environment to physical activity},
  author={Althoff, Tim and Ivanovic, Boris and King, Abby C and Hicks, Jennifer L and Delp, Scott L and Leskovec, Jure},
  journal={Nature},
  volume={645},
  number={8080},
  pages={407--413},
  year={2025},
  publisher={Nature Publishing Group UK London}
}

@article{dong2020interactive,
  title={An interactive web-based dashboard to track COVID-19 in real time},
  author={Dong, Ensheng and Du, Hongru and Gardner, Lauren},
  journal={The Lancet infectious diseases},
  volume={20},
  number={5},
  pages={533--534},
  year={2020},
  publisher={Elsevier}
}

@article{berger2021rational,
  title={Rational policymaking during a pandemic},
  author={Berger, Lo{\"\i}c and Berger, Nicolas and Bosetti, Valentina and Gilboa, Itzhak and Hansen, Lars Peter and Jarvis, Christopher and Marinacci, Massimo and Smith, Richard D},
  journal={Proceedings of the National Academy of Sciences},
  volume={118},
  number={4},
  pages={e2012704118},
  year={2021},
  publisher={National Academy of Sciences}
}

@inproceedings{zhao2024expel,
  title={Expel: Llm agents are experiential learners},
  author={Zhao, Andrew and Huang, Daniel and Xu, Quentin and Lin, Matthieu and Liu, Yong-Jin and Huang, Gao},
  booktitle={Proceedings of the AAAI Conference on Artificial Intelligence},
  volume={38},
  number={17},
  pages={19632--19642},
  year={2024}
}

@article{sahoo2024systematic,
  title={A systematic survey of prompt engineering in large language models: Techniques and applications},
  author={Sahoo, Pranab and Singh, Ayush Kumar and Saha, Sriparna and Jain, Vinija and Mondal, Samrat and Chadha, Aman},
  journal={arXiv preprint arXiv:2402.07927},
  year={2024}
}

@article{rahman2013survey,
  title={A survey on geographic load balancing based data center power management in the smart grid environment},
  author={Rahman, Ashikur and Liu, Xue and Kong, Fanxin},
  journal={IEEE Communications Surveys \& Tutorials},
  volume={16},
  number={1},
  pages={214--233},
  year={2013},
  publisher={IEEE}
}

@inproceedings{garrosi2017geo,
  title={Geo-routing in urban Vehicular Ad-hoc Networks: A literature review},
  author={Garrosi, Mehdi Tavakoli and Kalac, Marcel and Lorenzen, Torsten},
  booktitle={2017 International Conference on Computing, Networking and Communications (ICNC)},
  pages={865--871},
  year={2017},
  organization={IEEE}
}

@article{akhtar2012coordination,
  title={Coordination in humanitarian relief chains: chain coordinators},
  author={Akhtar, P and Marr, NE and Garnevska, EV},
  journal={Journal of humanitarian logistics and supply chain management},
  volume={2},
  number={1},
  pages={85--103},
  year={2012},
  publisher={Emerald Group Publishing Limited}
}

\newpage
\appendix

\section{Structured Prompts}
\label{appendix:prompt}
Here, an example prompt for 6-week TIR is provided.
\begin{tcolorbox}[
    enhanced,
    colback=gray!10!white, 
    colframe=gray!80!white, 
    colbacktitle=gray!70!white, 
    coltitle=white, 
    arc=2mm, 
    boxrule=1pt, 
    left=1mm,   
    right=1mm,  
    top=1mm,    
    bottom=1mm, 
]
\small 

\textbf{\# System Guidance:} \\
You are the epidemic control \& mobility policy assistant for state \emph{Arizona}. 
Your task is to recommend traffic control policies for this state to slow disease spread.  For each origin state, determine the inbound traffic allocation into \emph{Arizona} for the next 6 weeks. Total inbound Flow over 6 weeks must stay constant (equal to the ground truth); you may adjust weekly proportions.

\bigskip 

\textbf{\# Inputs:} \\
You will be given the following information for your task: 

- \textbf{State-level pandemic statistics}: The current population composition of the state \emph{Arizona} and its neighboring states, including  Susceptible (S), Exposed (E), Infected (I), Confirmed (Q), Recovered (R), Deaths (D)  

- \textbf{Historical inter-state mobility information}: Past \emph{21-day} average daily inbound flow (from other state to \emph{Arizona})

- \textbf{Planning-horizon mobility baselines}: average daily inbound flow for the upcoming \emph{42 days}

\bigskip 

\textbf{\# Constraints:}

You need to consider the following constraints:

\quad - output \emph{six} fractions for each origin state $i$, $[p_{i,1}, p_{i,2}, p_{i,3}, p_{i,4}, p_{i,5}, p_{i,6}]$ with $\sum_{m\in\{1,...,6\}}p_{i,m}=1$, and $p_{i,m}>0$ ; 
    
\quad - Week $m$ inbound flow from origin state $i$ = Total Flow * $p_{i,m}$
    
\quad - Low  $p_{i,m}$ means strict mobility control; high  $p_{i,m}$ means relaxed control

\bigskip 





\textbf{\# Final Output:} 

Organize your final answer as following:

\quad - \textbf{think\_process}: up to 200 words summary of reasoning process. 

\quad - \textbf{refined\_solution}: \{"state\_$i$": $[p_{i,1}, p_{i,2}, p_{i,3}, p_{i,4}, p_{i,5}, p_{i,6}]$,...\} 

\end{tcolorbox}

\section{Performance Comparison across LLM Models}
\label{appendix:llm_comparison}
\begin{figure}[!h]
  \centering
  \includegraphics[width=0.8\columnwidth]{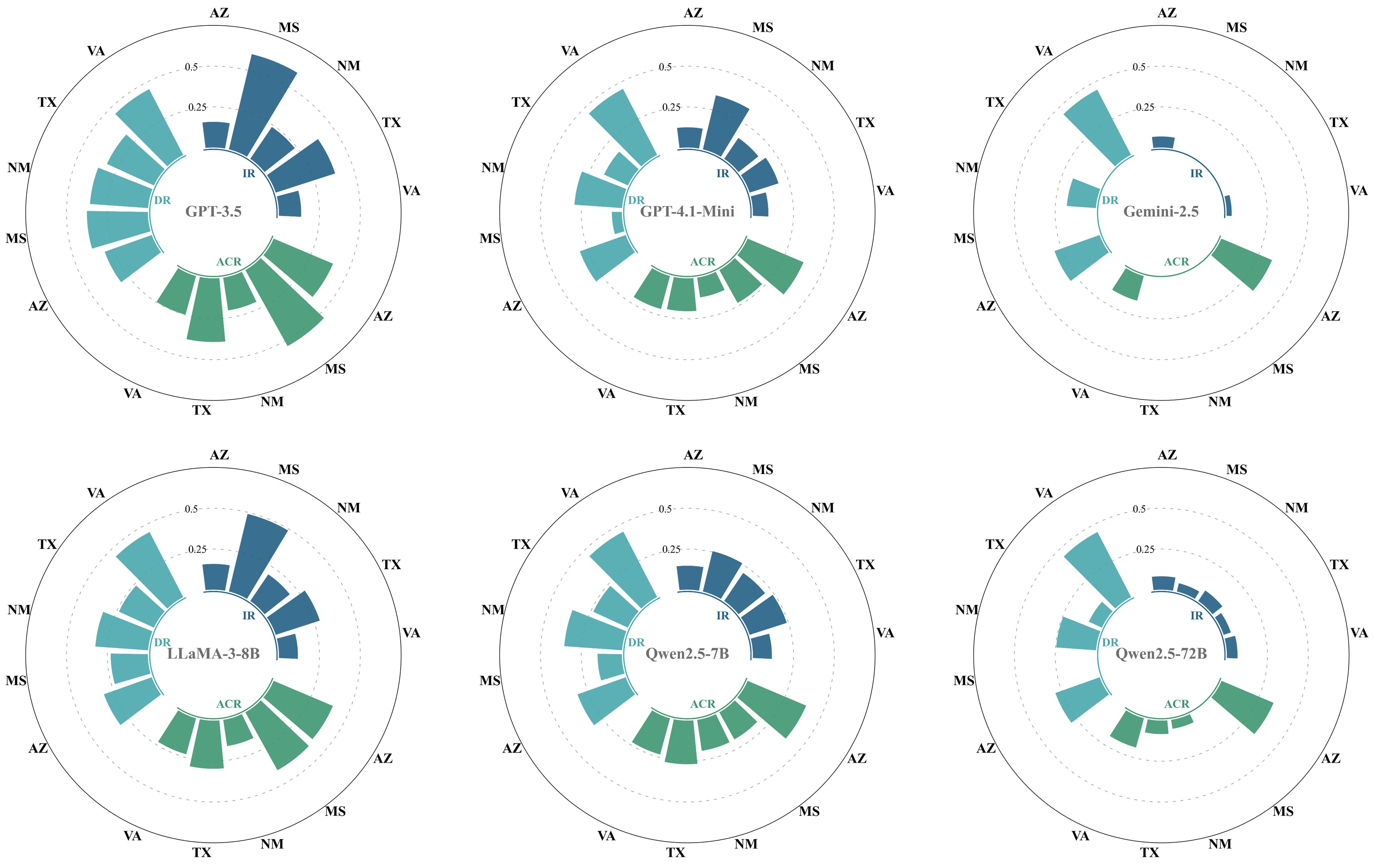}
  \caption{\textbf{Performance Comparison across LLM Models.} The proposed framework is tested with different LLM models, including GPT-3.5, GPT-4.1-Mini, Gemini-2.5, LLaMA-3-8B, Qwen2.5-7B, and Qwen2.5-72B. The improvement of DR, IR, and ACR, compared to the ground-truth policymaking in different states, is illustrated. GPT-3.5, GPT-4.1-Mini, LLaMA-3-8B, and Qwen-2.5-7B exhibit stable performance in pandemic control across states. }
  \label{fig:fig7}
  \vspace{-0.5cm}
\end{figure}

\newpage

\includepdf[pages=-]{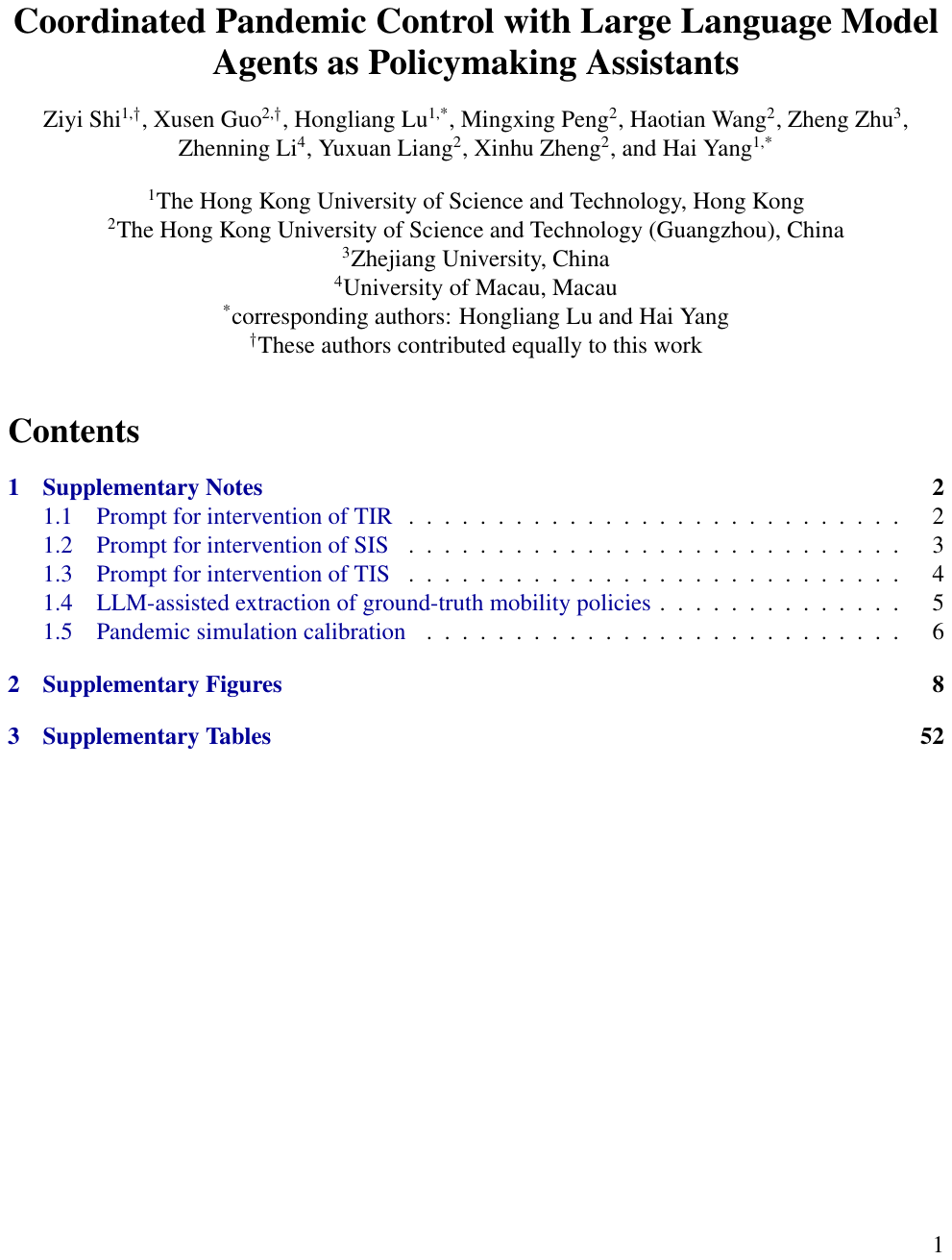}

\end{document}